\documentclass[conference, 10pt]{IEEEtran}
\IEEEoverridecommandlockouts
\usepackage{cite}
\usepackage{amsmath,amssymb,amsfonts}
\usepackage{textcomp}
\usepackage{kotex}
\usepackage{xurl}
\usepackage{adjustbox}
\def\BibTeX{{\rm B\kern-.05em{\sc i\kern-.025em b}\kern-.08em
    T\kern-.1667em\lower.7ex\hbox{E}\kern-.125em}}

\usepackage{amssymb}
\usepackage{xcolor}
\usepackage{graphicx}
\usepackage{mathtools,xparse}

\usepackage{setspace}
\usepackage{booktabs}

\usepackage{scalerel}

\usepackage{epstopdf}
\usepackage{xspace}
\newcommand{\Design}{FedUHD\xspace}

\usepackage{algorithm}% http://ctan.org/pkg/algorithms
\usepackage{algpseudocode}% http://ctan.org/pkg/algorithmicx
\algdef{SE}[DOWHILE]{Do}{doWhile}{\algorithmicdo}[1]{\algorithmicwhile\ #1}%

\usepackage{cprotect}
\usepackage{soul}
\usepackage{caption}
\usepackage{subcaption}

\usepackage{microtype}
\usepackage{makecell}
\usepackage{multirow}
\usepackage{multicol,tabularx,capt-of}
\usepackage{fancyvrb}

\usepackage{enumitem}
\setlist[itemize]{noitemsep, topsep=0pt, leftmargin=*}
\newcommand{\smallsqueezeup}{\vspace{-.6\baselineskip}}
\newcommand{\squeezeup}{\vspace{-\baselineskip}}
\usepackage{tikz}
\newcommand*\circled[1]{\tikz[baseline=(char.base)]{
            \node[shape=circle,fill,inner sep=0.5pt] (char) {\textcolor{white}{#1}};}}

\begin{document}
\title{\Design: Unsupervised Federated Learning using In-Memory Hyperdimensional Computing}

\author{
\IEEEauthorblockN{
You Hak Lee\textsuperscript{1}, 
Keming Fan\textsuperscript{1}, 
Xiaofan Yu\textsuperscript{2}, 
Quanling Zhao\textsuperscript{1}, 
Tianqi Zhang\textsuperscript{1}, 
Flavio Ponzina\textsuperscript{3}, 
and Tajana Rosing\textsuperscript{1}
}
\IEEEauthorblockA{
\textsuperscript{1}University of California San Diego,
\textsuperscript{2}University of California Merced, 
\textsuperscript{3}San Diego State University, USA\\
\{yhl004, k4fan, quzhao, tiz014, tajana\}@ucsd.edu, xiaofanyu@ucmerced.edu, fponzina@sdsu.edu
}
\thanks{\scriptsize \textcopyright~2026 IEEE. Personal use of this material is permitted. Permission from IEEE must be obtained for all other uses, in any current or future media, including reprinting/republishing this material for advertising or promotional purposes, creating new collective works, for resale or redistribution to servers or lists, or reuse of any copyrighted component of this work in other works. DOI: 10.1109/TCASAI.2026.3727571.}
\squeezeup
}

\maketitle

\begin{abstract}
Unsupervised federated learning (UFL) enables privacy-preserving distributed training without data labeling, yet practical deployment remains challenging due to non-IID data, high computational and communication costs at edge devices, and sensitivity to communication noise.
We propose \Design, the first UFL framework based on Hyperdimensional Computing (HDC). On the client side, \Design employs kNN-based cluster hypervector removal to mitigate non-IID effects by filtering detrimental local outliers. On the server side, cluster-aware HDC aggregation leverages cluster-level statistics to stabilize learning across heterogeneous clients.
To further improve efficiency, we design a compute-in-memory (CIM) accelerator based on a novel phase-change memory (PCM) device, integrated with lightweight ASIC digital modules to execute the client-side HDC pipeline within the accelerator. The intrinsic robustness of HDC to low precision and device variations enables efficient mapping onto analog PCM crossbars, exploiting massive parallelism while minimizing data movement.
Experimental results show that FedUHD achieves comparable accuracy to state-of-the-art neural network-based UFL methods across all datasets. 
On HAR and CIFAR10/100, FedUHD delivers an average 2,239$\times$ speedup and 1,542$\times$ higher energy efficiency on GPU.
In addition, \Design reduces communication cost by up to $176\times$ on HAR and CIFAR10/100 and demonstrates greater robustness than Orchestra under communication noise.
Compared to GPU implementation of \Design, the proposed PCM-based accelerator provides an additional $4.07\times$ speedup and three orders of magnitude higher energy efficiency on average. Furthermore, the results demonstrate the benefit of PCM over RRAM as a CIM substrate.

\end{abstract}

\section{Introduction}
Federated learning (FL) enables distributed model training across edge devices without sharing raw data, making it attractive for privacy-sensitive applications. However, supervised FL relies on labeled datasets, and obtaining high-quality annotations at scale is costly and labor-intensive. To address this limitation, unsupervised federated learning (UFL)~\cite{FedU, FedUL, FedX, Orchestra} has emerged as a promising direction that eliminates the need for labeled data.

\begin{figure}[t]
\centering
	  \includegraphics[width=.95\linewidth]{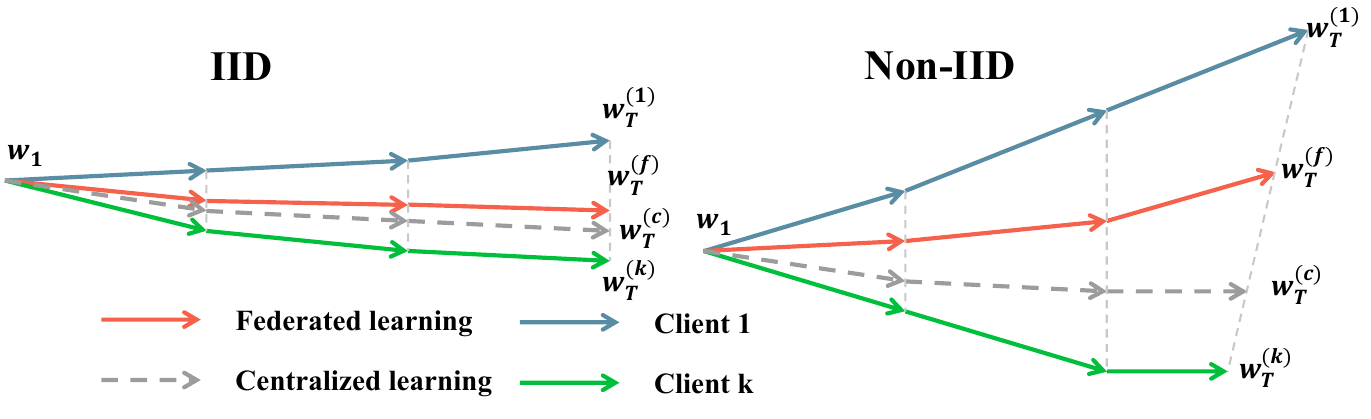}
  \caption{\small Visualization of weight divergence induced by non-IID data~\cite{FedU}. After $T$ rounds, non-IID data in federated learning causes substantial divergence from centralized learning  ($W_{T}^{(f)}$ vs.\ $W_{T}^{(c)}$).}
  \label{fig:noniid}
  \vspace{-1em}
\end{figure}

Despite its potential, practical UFL faces three fundamental challenges. 
(1) The presence of non-independent and non- identically distributed (non-IID) and unlabeled data across devices. In non-IID settings, globally aggregated updates can become unrepresentative or biased, leading to significant divergence from centralized training, as illustrated in Fig.~\ref{fig:noniid}. This issue is further exacerbated in unsupervised FL~\cite{FedU}. 
(2) High computational and communication costs at edge devices.
Training large NN models, such as the 11.7-million-parameter ResNet18~\cite{he2016deep}, requires costly backpropagation and model transfer, while communication-efficient methods such as FedPAQ~\cite{reisizadeh2020fedpaq} and SparsyFed~\cite{guastella2025sparsyfed} introduce additional quantization or sparse-indexing overhead on resource-constrained devices.
(3) Vulnerability to communication noise. Edge devices often operate in unreliable environments where packet loss and transmission noise are common~\cite{ye2022decentralized}. Since NN models rely on precise floating-point representations, bit-level errors during communication can cause exploding gradients or severe weight corruption, resulting in significant accuracy degradation or failure to converge~\cite{FedHD, FHDnn}.

Hyperdimensional Computing (HDC) provides a promising alternative to address these challenges. As a brain-inspired computational paradigm, HDC represents input data using high-dimensional vectors referred to as hypervectors (HVs). 
HDC-based model training is lightweight and does not require backpropagation, resulting in significantly lower computational cost~\cite{HygHD}. Moreover, its holistic representation ensures resilience to noise, as flipping individual HV elements does not substantially affect the encoded information~\cite{FHDnn}. 
HDC has been successfully applied to various machine learning tasks, including image classification~\cite{ge2020classification} and graph node classification~\cite{relHD, HygHD}, and has also been incorporated into supervised FL frameworks~\cite{FedHD, FHDnn}. In supervised FL settings, HDC achieves accuracy comparable to NN-based methods while reducing model size by over an order of magnitude and computational complexity by more than two orders of magnitude, making it well suited for edge deployment.
However, existing HDC-based FL approaches are confined to supervised learning~\cite{FedHD, FHDnn}. Although unsupervised HDC techniques have been explored for clustering~\cite{unsu_hdc} and lifelong learning~\cite{yu2024lifelong} on a single device, their extension to unsupervised federated environments remains largely unexplored. In the absence of labels, aligning locally learned prototypes across heterogeneous clients becomes a critical challenge. Naively aggregating misaligned representations results in destructive updates and unstable training.

In HDC-based learning, compute-in-memory (CIM) architectures are well suited to the inherent parallelism of HDC workloads. Since each HV element is typically accessed only once for a single operation, such as a multiply-accumulate (MAC) or bitwise computation, HDC exhibits low arithmetic intensity and becomes memory-bound in conventional architectures, often leading to suboptimal GPU utilization ($\approx$50\%). Prior work~\cite{HygHD, fan2024specpcm} has demonstrated that HDC can be efficiently mapped onto analog crossbar arrays. However, existing CIM accelerators have primarily focused on centralized learning workloads. In contrast, UFL with HDC introduces distinct algorithmic and dataflow requirements to address the challenges outlined earlier.

In this paper, we propose \Design, the first HDC-based framework for UFL, with a joint algorithm-hardware design tailored to the efficiency and robustness constraints of edge deployment. Our specific contributions are as follows:
\begin{itemize}
    \item \textbf{Algorithm}: We propose \Design, the first HDC-based framework for UFL, combining client-side kNN-based cluster hypervector removal with server-side cluster-aware HDC aggregation to mitigate non-IID unlabeled data and stabilize learning across heterogeneous clients.

    \item \textbf{Hardware}: We design a CIM accelerator based on a novel PCM device~\cite{pcm} with lightweight digital ASIC modules ($<$4\% area) to execute the client-side HDC pipeline. The architecture exploits analog crossbar parallelism and minimizes data movement, enabled by HDC’s robustness to low-precision computation and device variations.

    \item \textbf{Algorithm results:} Across all datasets, FedUHD achieves competitive accuracy against state-of-the-art NN-based UFL methods. On HAR and CIFAR10/100, it provides average 2,239$\times$/1,542$\times$ speedup and energy efficiency gains on GPU and 549$\times$/502$\times$ gains on CIM, while reducing communication cost by up to 176$\times$ and improving robustness under unreliable communication.
    \item \textbf{Hardware results}: The proposed PCM-based accelerator achieves $4.07\times$ / $2{,}981\times$ speedup and energy efficiency gains compared to GPU execution of the same \Design client-side pipeline, and outperforms an RRAM-based CIM baseline by $13.22\times$ / $6.04\times$ speedup and energy efficiency on average.
\end{itemize}

\section{Related Work and Background} 

\label{rel work}
\subsection{NN-based Unsupervised Federated Learning} \label{REL-NN-UFL}
Although early FL research has focused on supervised and semi-supervised scenarios, the scarcity of labeled data in practical deployments poses a fundamental barrier to their applicability~\cite{FedU}.
Recently, several studies~\cite{FedU, FedUL, FedX, Orchestra} have conducted research in UFL. 
FedU~\cite{FedU} tackled the degradation of local models due to data heterogeneity by implementing divergence-aware predictor updates. FedUL~\cite{FedUL} facilitated the transformation of UFL into a conventional supervised FL framework by introducing temporal labels called surrogate labels. FedX~\cite{FedX} improved the training efficiency and convergence stability in UFL by a two-sided knowledge distillation process between local and global models.
Orchestra~\cite{Orchestra} uses a clustering-based approach for UFL, maintaining local clusters on each client and global clusters in the cloud. However, Orchestra performs well only with easily clusterable representations.
The main issue with existing UFL research is its reliance on NN models.
NN models require significant resources to train~\cite{SAGE}, incur high communication costs, and also exhibit vulnerability to communication noise such as packet loss~\cite{FedHD, FHDnn}.

\begin{figure}[t]
\centering
	  \includegraphics[width=.95\linewidth]{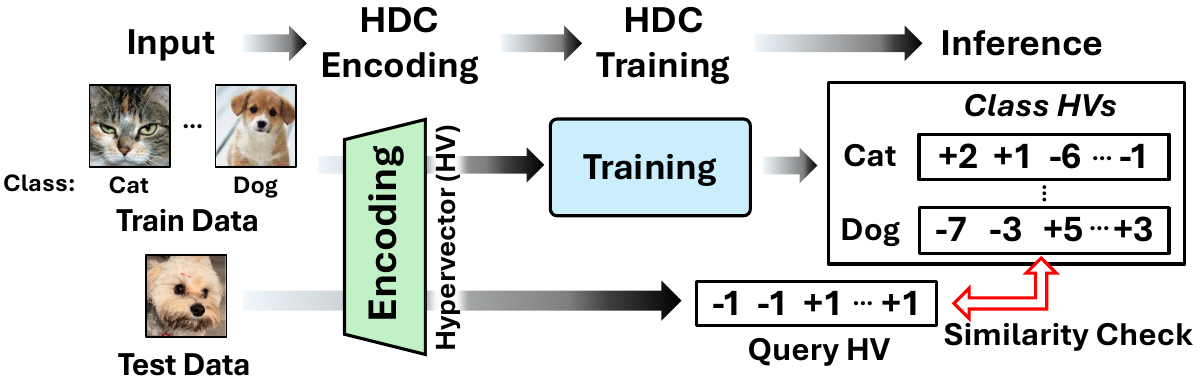}
  \caption{\small Overview of the HDC training and inference flow.
  }
  \label{fig:hdc}
  \vspace{-1em}
\end{figure}

\subsection{HDC Background} \label{PD-HDC-SFL}
Hyperdimensional Computing (HDC) represents data as high-dimensional, often low-precision ``hypervectors (HVs)'' and performs learning through simple, parallelizable operations such as element-wise addition and multiplication~\cite{thomas2021theoretical,imani2021revisiting}.
Its high-dimensional representation improves data separability, while its lightweight operations enable memory-efficient and fast training~\cite{khaleghi2021tiny}.

Fig.~\ref{fig:hdc} provides an overview of the HDC training and inference pipeline. The first stage of any HDC learning algorithm is encoding, which maps the data from its ambient representation to HVs. There are several approaches in HDC literature for encoding functions that preserve different types of similarity, such as Random Projection~\cite{rachkovskij2015formation} and Random Fourier Features~\cite{rahimi2007random}.  
After encoding, learning with HVs is straightforward: data HVs from the same category are bundled (e.g., via element-wise addition) to form a single class HV that aggregates class-specific information. Classification and inference are performed through a similarity search between query HVs and class HVs. Hamming distance is used when both query HVs and class HVs are binarized, while cosine similarity is used otherwise.

\subsection{HDC and Federated Learning}
Recent studies~\cite{FL-HDC, FHDnn, FedHD, RE-FHDC} have shown that HDC's benefits can be effectively applied to FL. Instead of transmitting model weights, lightweight and noise-tolerant class HVs learned on individual clients are exchanged, aggregated through bundling, and redistributed to the clients. This process is repeated for multiple rounds until the global model converges.

FL-HDC~\cite{FL-HDC} significantly reduces communication overhead by transmitting bipolarized model HVs instead of traditional high-precision model weights. 
RE-FHDC~\cite{RE-FHDC} further reduces computational and communication overhead by partitioning HVs during the training and transfer phases. 
FHDnn~\cite{FHDnn} integrates pretrained convolutional neural networks (CNNs) with HDC to combine the capability of DNNs with the efficiency of HDC in a federated setting. 
FedHD~\cite{FedHD} demonstrates HDC-based federated learning practically using real edge computing devices and under real network conditions. Despite the above contributions, these techniques are not applicable in scenarios without label supervision.

\begin{figure}[t]
\centering
	  \includegraphics[width=\linewidth]{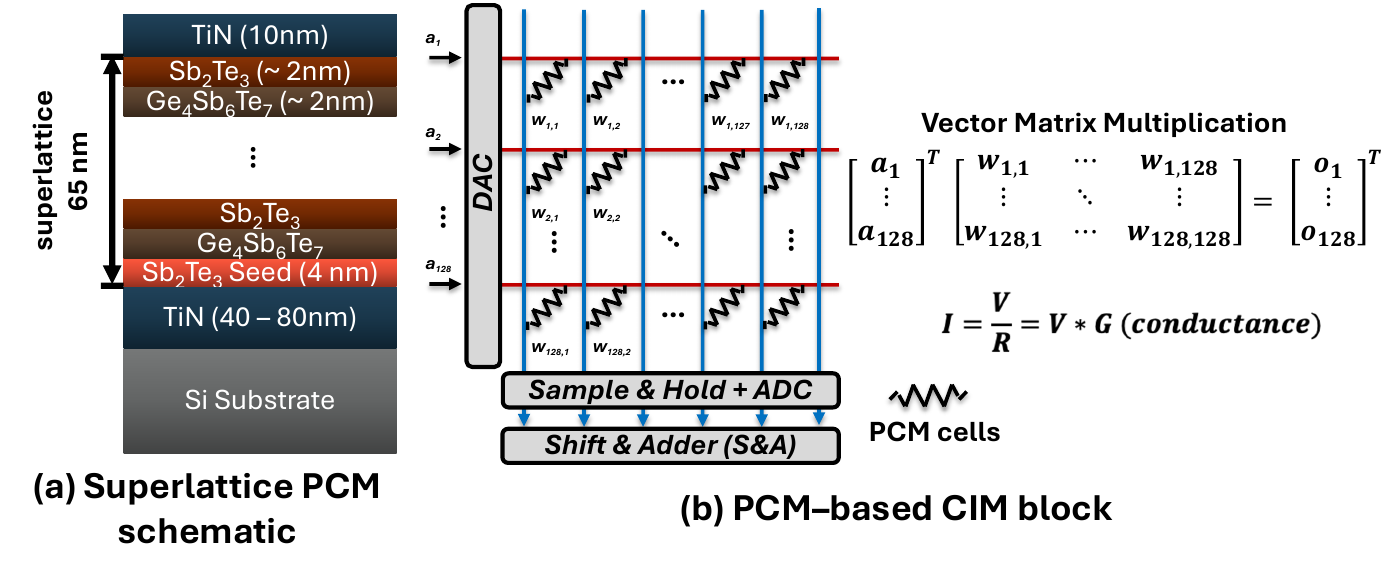}
  \caption{\small Superlattice PCM device~\cite{pcm} and PCM-based CIM block.
  }
  \label{fig:pcm}
  \vspace{-1em}
\end{figure}

\subsection{Compute-in-memory with PCM}
Compute-in-memory (CIM)~\cite{fan2024specpcm, chi2016prime, fan2024efficient, yoon202129} has emerged as a promising solution to mitigate the overhead of frequent data movement between memory and processors in conventional CPU/GPU systems by enabling computation directly where data reside.
Compared with digital CIM, analog CIM can deliver over two orders of magnitude higher energy efficiency~\cite{fan2024efficient} by exploiting low-voltage-swing analog operations, particularly for multiply-and-accumulate (MAC). Among memory technologies for analog CIM, RRAM has been widely considered due to its fast read access, simple fabrication, and suitability for multi-level cell operation. However, its  high write latency and high programming voltage ($>$2V) limit its efficiency for update-intensive workloads such as training and clustering~\cite{fan2024efficient}.

Recent advances in phase-change memory (PCM) introduce superlattice structures (e.g., Sb\(_2\)Te\(_3\)/Ge\(_4\)Sb\(_6\)Te\(_7\))~\cite{pcm} (see Fig.~\ref{fig:pcm}(a)) that offer low programming voltage ($<$1V), low switching energy, reduced resistance drift, and strong process maturity.
These properties support efficient multi-level storage and good compatibility with CMOS logic~\cite{pcm}, making PCM a compelling substrate for analog CIM accelerators. 

In a typical PCM-based CIM block as shown in Fig.~\ref{fig:pcm}(b), memory cells are organized in a crossbar array, with each cell storing a programmable conductance $w_i$.
Inputs are applied as voltages $a_i$ to the rows, generating currents that are summed along the columns to produce the outputs, enabling fast and energy-efficient computation directly within memory.
While PCM-based CIM provides substantial benefits, it also exhibits non-idealities such as resistance drift, process variation, and voltage fluctuations. 
HDC's intrinsic error resilience naturally helps mitigate these effects in PCM-based CIM systems.

\section{FedUHD Algorithm}
FedUHD is the first HDC-based UFL framework that is more energy efficient and robust than NN-based methods, while removing the reliance on labeled data in comparison to prior HDC-based supervised FL. 
FedUHD inherits the lightweight operations and small model sizes of HDC, which enable high computational and communication efficiency.

However, a key challenge in HDC-based UFL is effectively training local models with unlabeled non-IID data on local devices and aggregating these models at a central server. 
The absence of class label information during local training and global aggregation makes it difficult to construct and combine models. This challenge is further exacerbated under non-IID data distributions, where discrepancies between local and global models become more pronounced (see Fig.~\ref{fig:noniid}). 
Model convergence is increasingly challenging under non-IID data distributions without labels~\cite{FedU}.
A global model aggregated on the server may be misaligned with a client’s local data distribution. Blindly using the global model during local training can introduce conflicting updates and amplify divergence between local and global models. Similarly, naively aggregating local models across clients without accounting for distribution mismatch can further hinder convergence (details are in~\ref{KCHR} and~\ref{CAHA}). 

The proposed \Design introduces two new techniques to address these challenges. First, on the client side, we propose a kNN-based cluster HV removal technique to accept only partial global information that is beneficial for training on each local device. Second, on the server side, we design a cluster-aware HDC aggregation approach to effectively balance the non-IID data distribution among clients even without labels (see the ablation study in Fig.~\ref{fig:ablation}).

\begin{figure}[t]
\centering
	  \includegraphics[width=\linewidth]{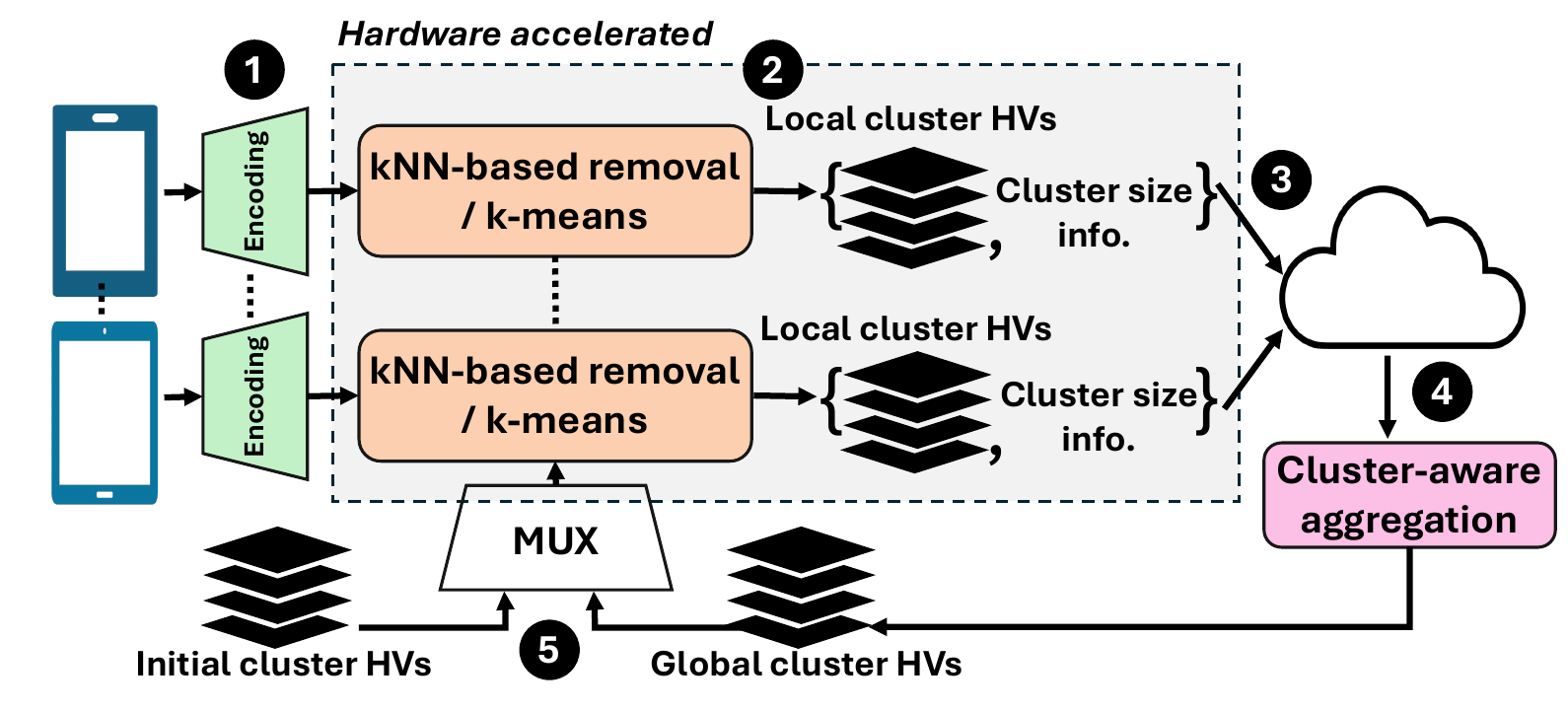}
  \caption{\small \Design framework overview.}
  \label{fig:overview}
  \vspace{-1em}
\end{figure}

\begin{algorithm}[t]
\caption{FedUHD: Modular Training Procedure}
\label{alg:feduhd}
\begin{algorithmic}[1]

\Procedure{ClientUpdate}{$\mathcal{X}_{i}, \mathcal{G}, rnd$}
    \State \textit{// HDC encoding \& centroid (cluster) initialization}
    \If{$rnd = 0$}
        \State $\mathcal{H} \leftarrow \Call{HDC\_Encode}{\mathcal{X}_{i}}$
        \State $\mathcal{G}_{init} \leftarrow \mathcal{O}$ \Comment{Random global cluster HVs}
    \Else
        \State \textit{// kNN-based cluster HV removal}
        \State $\mathcal{G}_{init} \leftarrow \{ \mathbf{g} \in \mathcal{G} \mid \text{has\_same\_id\_neighbor}(\mathbf{g}, \mathcal{H}) \}$ %\Comment{Filter relevant centroids}
        %\If{$\mathcal{G}_{init} = \emptyset$} $\mathcal{G}_{init} \leftarrow \mathcal{O}$ \EndIf
    \EndIf
    
    \State \textit{// Clustering and 3-bit centroid quantization}
    \State $\mathcal{L}_{i} \leftarrow 
    \text{quant\_3bit}(
    \text{kMeans}(\mathcal{H}, \text{init}=\mathcal{G}_{init}, \text{iter}=E))$
    
    \State \textbf{Return} ($\mathcal{L}_{i}$, $S_i$) \Comment{Local cluster HVs and size info.}
\EndProcedure
\Statex \hrulefill
\Procedure{ServerAgg}{$\{(\mathcal{L}_i, S_i)\}_{i=1}^{I}$}
    \State Initialize updated global set $\mathcal{G}_{new} = \emptyset$
    \State \textit{// Cluster-aware HDC aggregation}
    \For{$j = 1$ \textbf{to} $J$}
        \State $W_{ij} \leftarrow S_{ij} / \sum_{q=1}^{I} S_{qj}$ \Comment{Compute cluster weights}
        \State $\mathbf{g}_j \leftarrow \sum_{i=1}^{I} W_{ij} \cdot \mathbf{l}_{ij}$ \Comment{Weighted aggregation}
        \State Add $\mathbf{g}_j$ to $\mathcal{G}_{new}$
    \EndFor
    \State \textbf{Return} $\mathcal{G}_{new}$
\EndProcedure
\end{algorithmic}
\end{algorithm}

\subsection{Overview of \Design}
The objective of \Design is to train a set of global cluster HVs, $\mathcal{G} = \{\mathbf{g}_j \in \mathbb{R}^{D} \}$, without ground-truth labels, where $j \in \{1, ..., J\}$ and $D$ is the HD dimension. In unsupervised learning, the exact number of classes (labels) $K$ is unknown, so we assume a larger number of centroids or clusters, denoted $J$ ($J>K$). The total number of clients is $I$. $\mathcal{X}_i= \{\mathbf{x}_p \}_{p=1}^{n_i}$ represents the local training dataset for the client $i$.
Let $n_i=|\mathcal{X}_i|$ denote the number of data samples of client $i$. 

Formally, \Design approximately optimizes the global objective $\min_{\{\mathbf{g}_j\}} \sum_i\sum_{\mathbf{h}\in\mathcal{H}_i} \min_j d_{\mathrm{HDC}}(\mathbf{h},\mathbf{g}_j)$, where $\mathcal{H}_i$ is client $i$'s encoded dataset and $d_{\mathrm{HDC}}(\mathbf{h},\mathbf{g})=-\mathbf{h}^{\top}\mathbf{g}$ is the negative dot-product distance.
With the support mask $m_{ij}\in\{0,1\}$ estimated by kNN-based removal, client $i$ approximately solves $\min_{\{\mathbf{l}_{ij}:m_{ij}=1\}} \sum_{\mathbf{h}\in\mathcal{H}_i} \min_{j:\,m_{ij}=1} d_{\mathrm{HDC}}(\mathbf{h},\mathbf{l}_{ij})$, where $\mathbf{l}_{ij}$ is its local cluster HV.
The server updates $\mathbf{g}_j=\sum_i(S_{ij}/\sum_{q=1}^I S_{qj})\,\mathbf{l}_{ij}$, where $S_{ij}$ is the size of local cluster $j$.
Intuitively, the mask retains locally supported clusters, local training constructs their cluster HVs, and size-weighted aggregation updates the global cluster HVs to represent all clients' data.

To train global cluster HVs, \Design consists of 5 steps as shown in Fig.~\ref{fig:overview} and Algorithm~\ref{alg:feduhd}. Training is mainly divided into local training (steps~\circled{1}-\circled{3}) and global aggregation (steps~\circled{4}-\circled{5}).

\circled{1} \textbf{HDC encoding} maps client data to a high-dimensional space, producing data HVs. \Design uses standard Random Projection~\cite{thomas2021theoretical} for simple (e.g., non-image) data or incorporates a shallow pretrained feature extractor~\cite{he2016deep, marcel2010torchvision} with the projection to handle image input~\cite{dutta2022hdnn}. 
Using a pretrained feature extractor is a common practice in HDC pipelines for image processing~\cite{FHDnn, FedHD}, as it produces semantically meaningful representations that improve overall accuracy relative to directly applying Random Projection to raw image pixels.
The feature extractor remains frozen during both training and inference, ensuring a good balance between accuracy and energy efficiency~\cite{dutta2022hdnn}. Accordingly, we binarize the projected vectors to obtain the final data HVs.

Once all training data are converted to data HVs, the HDC encoding step is completed. We refer to the list of encoded data HVs as $\mathcal{H}$.

\circled{2} \textbf{Local cluster HV training} constructs local \textbf{cluster} HVs ($\mathcal{L}$), which serve the same
representational role as class HVs in supervised HDC, but are learned directly from data without labels due to the unsupervised
FL setting.
Unlike existing distributed clustering methods~\cite{balcan2013distributed} or Orchestra~\cite{Orchestra}, \Design performs local k-means clustering on encoded HVs ($\mathcal{H}$), offering better separability and robustness due to the high dimensionality.
Specifically, all clients begin with shared initial cluster HVs ($\mathcal{O}$) randomly generated on the server, each having a unique \textit{cluster\_id}, so that cluster HVs remain aligned across clients and can be aggregated meaningfully in the absence of labels. 
%The high-dimensional data points to be clustered are in $\mathcal{H}$ from the HDC encoding step. 
The number of iterations of local clustering is set to the number of local epochs $E$.

After the initial local clustering, \Design applies the proposed kNN-based cluster HV removal process (elaborated in Section~\ref{KCHR}) to remove outlier global/local cluster HVs resulting from the non-IID data distribution.
Subsequently, \Design conducts local clustering again with the remaining global cluster HVs as the new initial centroids. 
The number of local clusters is determined by the size of the initial centroid set $G_{\mathrm{init}}$ used in k-means. $G_{\mathrm{init}}$ contains $J$ shared cluster HVs in the first round and the remaining global cluster HVs after kNN-based removal in subsequent rounds.

$k_n$ denotes the number of nearest neighbors used in the kNN-based removal process and controls the strength of cluster HV removal. In kNN and k-means, the negative dot product is adopted as the distance metric, which aligns well with the MAC operations supported by CIM blocks and remains compatible with our mixed-precision representation of binary data HVs and 3-bit centroid HVs. 
Unlike data HVs, centroids are averages of multiple data HVs, so binarizing them discards substantial information and degrades accuracy. We therefore use the maximum 3-bit precision supported by our PCM-CIM to preserve centroid information.

\circled{3} \textbf{Local model transfer} transmits each
client's local model to the server. Unlike NN-based UFL, \Design represents the model as a set of local cluster HVs ($\mathcal{L}_i$). In addition to the cluster HVs, \Design sends the size of each local cluster ($S_{ij}$) to the server. This value is obtained by counting the $\textit{cluster\_id}$ assigned to each encoded sample. The cluster size information is required during global aggregation to properly weight local cluster HVs. The communication overhead is minimal since it consists of a single integer per cluster.

\circled{4} \textbf{Global aggregation} constructs the global
model ($\mathcal{G}$) as a set of global cluster HVs. Each local model captures data representations specific to an individual client and therefore lacks generality. To construct a global model that reflects information from all clients, these local models must be aggregated at the server. However, in the unsupervised setting, the absence of label information makes direct aggregation ill-defined. To address this challenge, \Design initializes all clients with a shared set of cluster HVs generated at the server, each carrying a unique \textit{cluster\_id} introduced in Step~\circled{2}. This persistent cluster identity preserves alignment across clients and enables coherent aggregation of local cluster HVs without supervision. In addition, the contribution of each local model can vary significantly across clients. To resolve these issues, we propose a cluster-aware HDC aggregation scheme detailed in Section~\ref{CAHA}.

\circled{5} \textbf{Global model transfer} delivers the global
model from the server to the clients. Before the start of the first communication round ($rnd$=0), a set of initial cluster HVs ($\mathcal{O}$) is randomly generated and sent to each client. In subsequent rounds, the server sends the latest global model $\mathcal{G}$ resulting from global aggregation.

\begin{figure}[t]
\centering
	  \includegraphics[width=.98\linewidth]{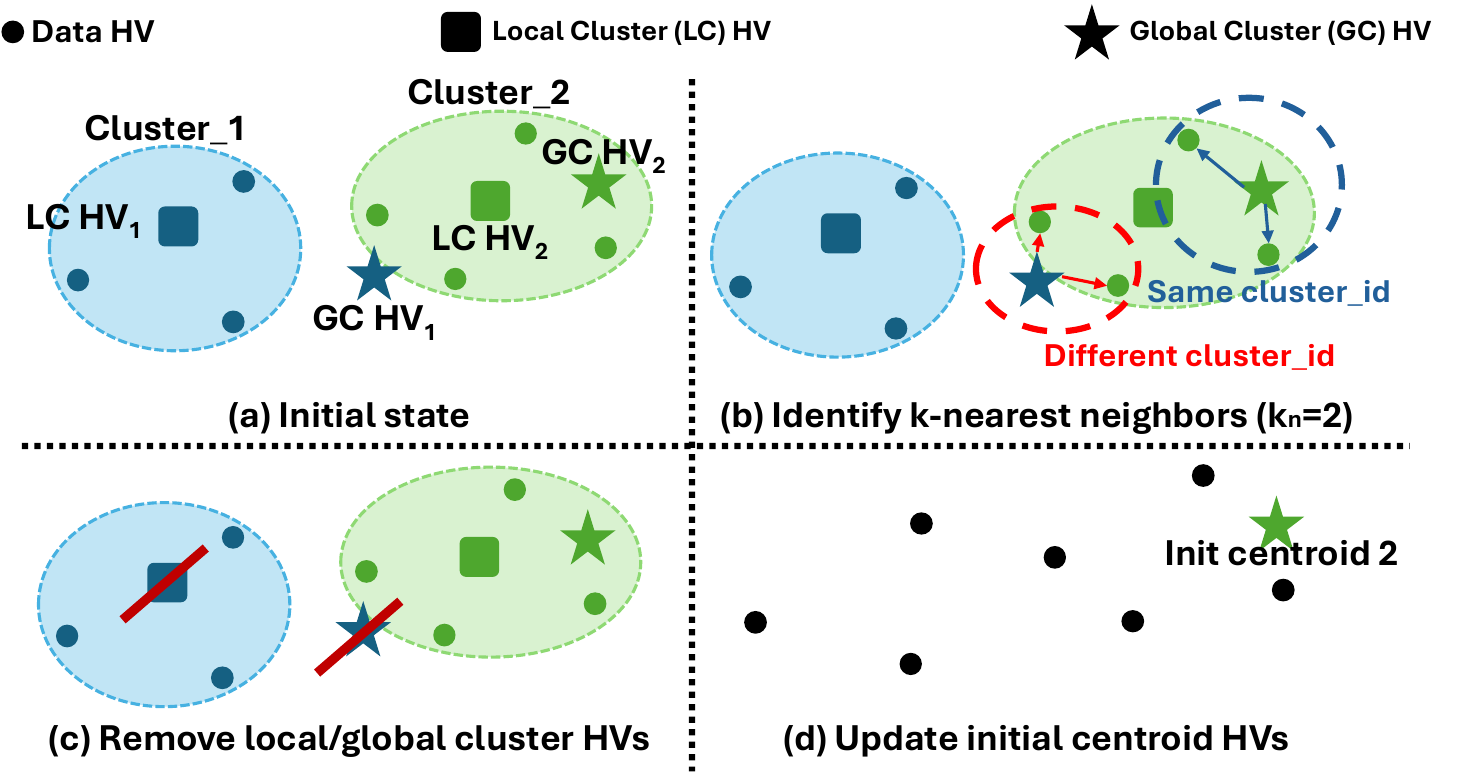}
  \caption{\small Visualization of the kNN-based cluster HV removal process.}
  \label{fig:knn_removal}
  \vspace{-1em}
\end{figure}

\subsection{kNN-based Cluster HV Removal on Clients} \label{KCHR}
In non-IID scenarios, local models may become biased toward specific distributions and the aggregated global model misrepresents these local distributions. Consequently, when the global model is sent back to local devices, it may hinder rather than facilitate model convergence. To this end, we introduce a kNN-based cluster HV removal technique for \Design. We integrate a kNN-based filtering approach rather than a k-means-only approach since k-means does not guarantee the identification and removal of global cluster HVs that are irrelevant to a client’s local data distribution in non-IID settings, while kNN enables explicit detection of cluster HVs lacking local neighborhood relevance.

Fig.~\ref{fig:knn_removal} shows the detailed process of kNN-based cluster HV removal in a simple example. The removal consists of four stages: initial state, identify k-nearest neighbors, remove local and global cluster HVs, and update initial centroid HVs for the current round. Fig.~\ref{fig:knn_removal}~(a) illustrates the state before applying the kNN-based cluster HV removal. As a result of the previous round of local training, two clusters have been formed, with each color representing one \textit{cluster\_id}. The individual points represent the data HVs, while the squares denote the local cluster HVs from the previous round, and the stars indicate the global cluster HVs received from the server after the global aggregation.

In the ``identify k-nearest neighbors'' phase (Fig.~\ref{fig:knn_removal} (b)), each global cluster HV is examined by identifying the \textit{cluster\_id}s of its $k_n$ nearest neighboring data HVs, where $k_n$ is a predefined value. 
Fig.~\ref{fig:knn_removal}~(b) illustrates a case where $k_n$ is set to 2. 
For both Global Cluster HV 1 ($\mathbf{g}_1$) and Global Cluster HV 2 ($\mathbf{g}_2$), the nearest data HVs have \textit{cluster\_id} 2. Since $\mathbf{g}_1$ has \textit{cluster\_id} 1, it has no same-ID neighbor and is removed, whereas $\mathbf{g}_2$, whose \textit{cluster\_id} is 2, is retained.
If, during the ``identify k-nearest neighbors'' phase, a global cluster HV has \textbf{no} neighboring data HVs with the same \textit{cluster\_id}, \Design treats the global cluster HV as an ``outlier'' and the corresponding global and local cluster HVs of that \textit{cluster\_id} are removed in the ``remove local and global HVs'' stage for the current round. Fig.~\ref{fig:knn_removal}~(c) shows the removal of Global/Local Cluster HV 1.
This step filters out the \textit{cluster\_id} only from the current client's local clustering, rather than from the global model itself.
Finally, the initial centroid HVs for the current round's local clustering are updated using only the remaining global cluster HVs (Fig.~\ref{fig:knn_removal}~(d)). Equivalently, the filtered set $G_{\mathrm{init}}$ in Alg.~\ref{alg:feduhd} contains only the \textit{cluster\_id}s that pass the kNN-based removal, and k-means is performed using these remaining centroids as initialization.

\subsection{Cluster-aware HDC Aggregation in Cloud} \label{CAHA}
To aggregate local models under label-free and non-IID data distributions, we propose a cluster-aware HDC aggregation method executed in the cloud. Unlike supervised HDC-based FL, the absence of labels prevents the direct application of aggregation techniques such as bundling~\cite{FHDnn} or FedAvg~\cite{mcmahan2017communication}, which rely on consistent class correspondence across clients. In unsupervised settings, clients may learn different numbers of local clusters. For example, one client may produce $m$ local clusters while another produces $n$, making it unclear how local clusters should be matched and combined during global aggregation. To resolve this ambiguity, \Design utilizes the \textit{cluster\_id} established during local initialization to enforce a unified structural reference. By mapping each local cluster HV to a globally synchronized \textit{cluster\_id}, the framework facilitates coherent alignment and persistent aggregation across successive communication rounds without labels. Furthermore, recognizing that client contributions to specific clusters vary, our approach integrates a cluster-aware aggregation scheme for local cluster HVs. As illustrated by the comparative analysis in Fig.~\ref{fig:ablation}, failing to account for these distributional differences leads to stochastic instability and a marked reduction in model efficacy.

Let $\mathbf{l}_{ij}$ denote the local cluster HV on client $i$ with cluster\_id $j$, and let $S_{ij}$ denote the size of cluster $j$ on client $i$.
\Design performs cluster-aware HDC aggregation to combine the local models and obtain a new global cluster $\mathbf{g}_j$ as follows:
\begin{equation}
\begin{split}
W_{ij} = \frac{S_{ij}}{\sum_{q=1}^{I}{S_{qj}}}, 
\mathbf{g}_j = \sum_{i=1}^{I}{W_{ij}*\mathbf{l}_{ij}} \\
\end{split}
\end{equation}
\\
Even when clients produce different numbers of local clusters, \Design remains well defined since aggregation is performed independently for each \textit{cluster\_id}. For a given \textit{cluster\_id} $j$, only clients that locally instantiate this cluster contribute to the aggregation, while clients without this cluster implicitly have $S_{ij}=0$ and thus receive zero weight. As a result, missing clusters on some clients do not affect aggregation, and the server aggregates only relevant local cluster HVs, yielding a balanced global model in non-IID data scenarios without requiring labels or assuming identical class presence.

\begin{figure}[t]
\centering
	  \includegraphics[width=\linewidth]{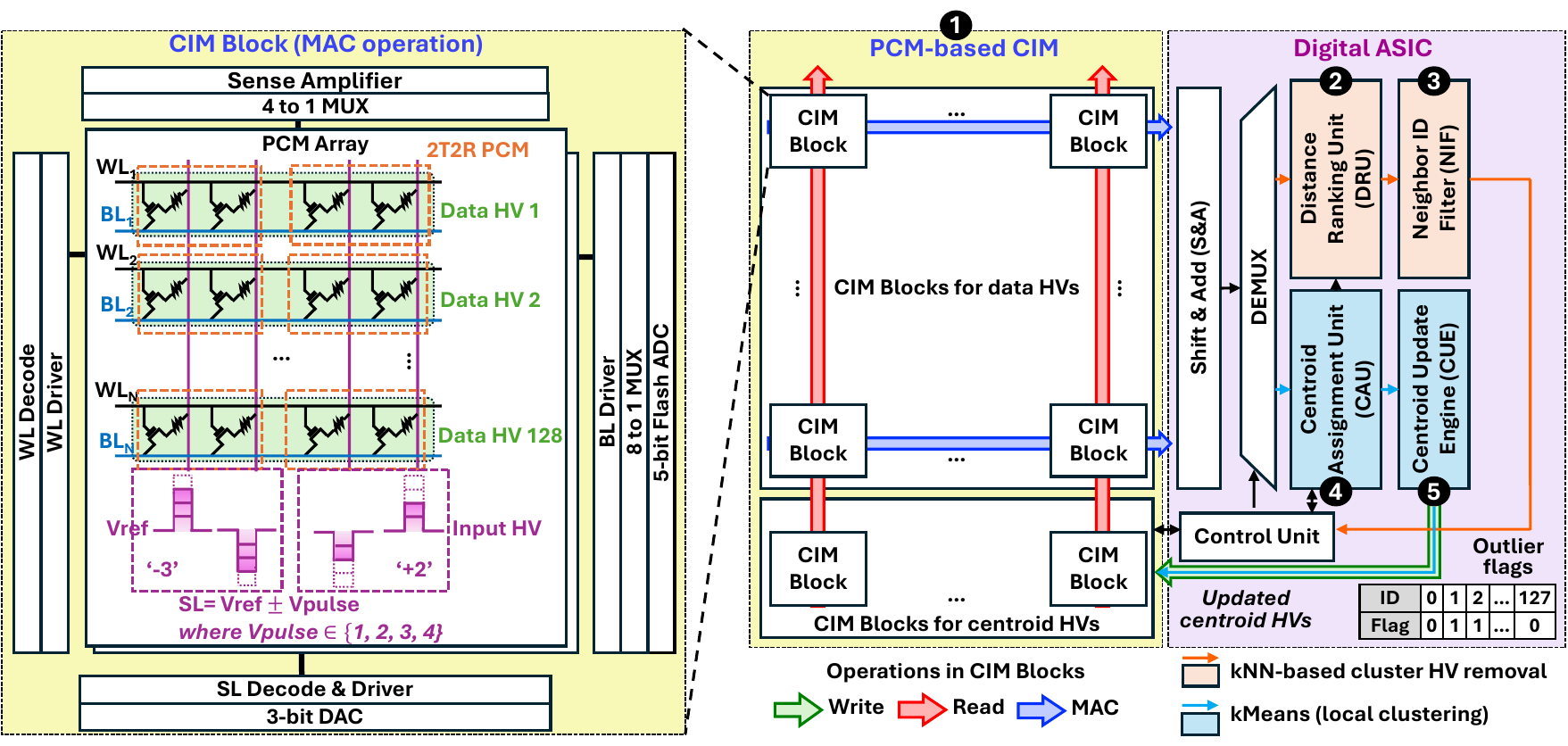}
    \caption{\small FedUHD accelerator architecture overview. The left panel illustrates a CIM block MAC operation: data HV segments are stored in the 2T2R PCM array, and 3-bit input elements are applied through the source line (SL) as $V_{\mathrm{ref}}\pm V_{\mathrm{pulse}}$, producing bitline currents that are digitized by row-multiplexed ADCs.}
  \label{fig:cim-overview}
  \vspace{-1em}
\end{figure}

\section{FedUHD Accelerator}
We design an accelerator for \Design by leveraging the benefits of CIM in newer superlattice PCM memory~\cite{pcm} to accelerate the distance computation, which accounts for 90\% of the total operations.  We also integrate lightweight ASIC digital modules to execute the remaining non-CIM-friendly operations such as ranking/selection, filtering, and quantization, as well as to orchestrate client-side execution. 
Recent superlattice PCM~\cite{pcm} improves programming voltage/energy, resistance window, and drift characteristics while remaining CMOS-compatible. Compared with RRAM, these properties reduce programming overhead and help mitigate variability and read/programming noise in analog CIM. Compared with SRAM, PCM is non-volatile and provides denser storage for the large signed HVs, since an SRAM is based on a 6T structure.
However, practical PCM-CIM must still contend with (i) non-idealities such as resistance drift, process variation, and voltage fluctuations, and (ii) read/write asymmetry where writes remain more expensive than reads. To address these issues, we leverage HDC's tolerance to low-precision arithmetic and its inherent error resilience to enable reliable execution on noisy PCM-CIM~\cite{FHDnn}. 

Fig.~\ref{fig:cim-overview} provides an overview of the proposed FedUHD client-side accelerator which consists of five modules: \circled{1}-\circled{5}. PCM-CIM \circled{1} exploits HDC's abundant data-level parallelism through in-memory vector operations to perform distance computations between data and centroid hypervectors, thus reducing off-chip traffic. While prior work has shown that HDC maps efficiently to analog crossbar-based CIM~\cite{HygHD, fan2024specpcm}, most existing designs target centralized learning. UFL with HDC introduces distinct algorithmic and dataflow requirements that are not addressed by previous accelerators. \Design mapping and CIM operations are detailed in Sec.~\ref{cim_ops}. 

The digital ASIC modules \circled{2}-\circled{5} take the computed distances from PCM-CIM and complete the client-side clustering dataflow. Specifically, \circled{2} (Distance Ranking Unit, DRU) and \circled{3} (Neighbor ID Filter, NIF) support the kNN-based cluster HV removal and are invoked once at the beginning of each communication round to filter outlier global centroids. In contrast, \circled{4} (Centroid Assignment Unit, CAU) and \circled{5} (Centroid Update Engine, CUE) support the local k-means procedure and are executed at every local iteration to update local centroid HVs. We detail the design and operation of the digital ASIC modules in Sec.~\ref{digital_ops}.

\subsection{PCM-based CIM and Data Mapping}
\label{cim_ops}
The \Design accelerator executes high-dimensional element-wise operations directly inside memory to minimize data movement cost. Fig.~\ref{fig:cim-overview} illustrates the CIM block used in the \Design accelerator. 
Our design is fully homogeneous: the same CIM blocks are reused for both computation and storage. When configured to process data HVs, they operate in CIM mode to perform in-memory distance calculations. When configured for centroid HVs, the identical arrays operate in storage mode, serving only for read/write updates and query HV provisioning.
Each array contains a 128×128 grid of 2T2R cells. For clarity, we refer to the row and column indices of a cell within a PCM array as the \textit{cell row} and \textit{cell column}. In this structure, the PCM devices act as resistive elements,
and each 2T2R pair encodes a 3-bit signed value by exploiting the conductance difference between the two paired PCM devices. We model the 128$\times$128 CIM array by replicating this differential 2T2R cell into a standard WL/BL/SL array organization. The device-level resistance-state and programming characteristics are taken from the fabricated superlattice PCM device in~\cite{pcm}, while the access-device footprint and array/peripheral area are conservatively estimated at the circuit level.
% and every cell encodes a 3-bit signed value by relying on the conductance difference between its paired PCM elements.
Although storing 3 bits per cell introduces larger device- and circuit-level errors than a single-level cell~\cite{pcm}, the \Design algorithm exhibits strong tolerance to such noise, which enables effective use of the PCM cell properties (see Fig.~\ref{fig:algo_analysis}(a)).

The CIM blocks are organized in an $M\times N$ layout, as shown in Fig.~\ref{fig:cim-overview}, and we refer to the row and column positions of a CIM block within this grid as the \textit{}{array row} and \textit{array column}. Because the dimensionality of an encoded HV typically ranges from 1k to 10k~\cite{FedHD, relHD}, multiple CIM blocks are required to store a single HV. Each HV is partitioned into segments of 128 elements, which is the number of \textit{cell columns} in an array. 
Since each 2T2R cell supports up to a 3-bit signed value, we use this maximum precision for centroid HVs to preserve their averaged information, while data HVs remain binary. All elements belonging to the same HV are mapped to the same \textit{cell row} address.
Thus, each 128-element HV segment is mapped along the \textit{cell columns} of a 128$\times$128 2T2R PCM array.
For instance, an HV with dimension 1024 is divided into eight segments (=1024/128), requiring eight CIM blocks located on the same \textit{array row}. Each segment is written to the same \textit{cell row} across these arrays. In this layout, each \textit{cell row} across the array grid corresponds to a different HV, and each \textit{cell column} corresponds to the same element position within a segment. This mapping allows all CIM blocks along an \textit{array row} to operate in parallel, which maximizes throughput during vector-matrix operations.

\noindent 
\textbf{Operations in CIM block}: The CIM blocks in \Design support three fundamental operations: write, read, and multiply-and-accumulate (MAC). During a write operation, the word-line (WL) decoder activates the selected row while the bit-lines (BLs) are grounded. The digital HV elements are translated into corresponding analog amplitudes through 3-bit digital-to-analog converters (DACs). These programming pulses are delivered to the selected source-lines (SLs) via SL drivers, thereby programming the conductance levels of the PCM devices.  For a read operation, the WL decoder selects the target row to enable all 2T2R cells along that row. A small non-destructive voltage ($<$ 0.4 V) is applied to the BLs, and the resulting currents are sensed and amplified to reconstruct the stored HV in digital form.

During the MAC operation, one operand vector is first converted into analog voltages using DACs. All WLs are asserted simultaneously, applying the same input across all rows of the array. As illustrated in Fig.~\ref{fig:cim-overview}, the SL driver generates the analog input voltage from the digital input. Each 3-bit (8-level) element is mapped to $V_{\mathrm{ref}} \pm V_{\mathrm{pulse}}$, where $V_{\mathrm{pulse}} \in \{1,2,3,4\}$. The example shows the corresponding mappings for input elements of $-3$ and $+2$. The multiplication between the input voltage and the stored PCM conductance is performed intrinsically within each cell, and the resulting currents accumulate along the BLs. 
These accumulated currents are digitized by row-multiplexed 5-bit flash ADCs to produce partial dot products. Each 128$\times$128 array uses 16 ADCs in total, with each ADC shared across 8 rows through an 8:1 multiplexer. 
The partial dot products from different 128-element HV segments are then combined by shift \& add (S\&A) modules to form the final dot-product value.
To reduce area and power overhead, we adopt this row-multiplexed ADC organization rather than dedicating one ADC per row.
Under this design, the 16 row-multiplexed 5-bit flash ADCs contribute 25\% of the area overhead.
A detailed area breakdown and further analysis are provided in Sec.~\ref{CIM_analysis}.
 
\begin{figure}[t]
\centering
	  \includegraphics[width=\linewidth]{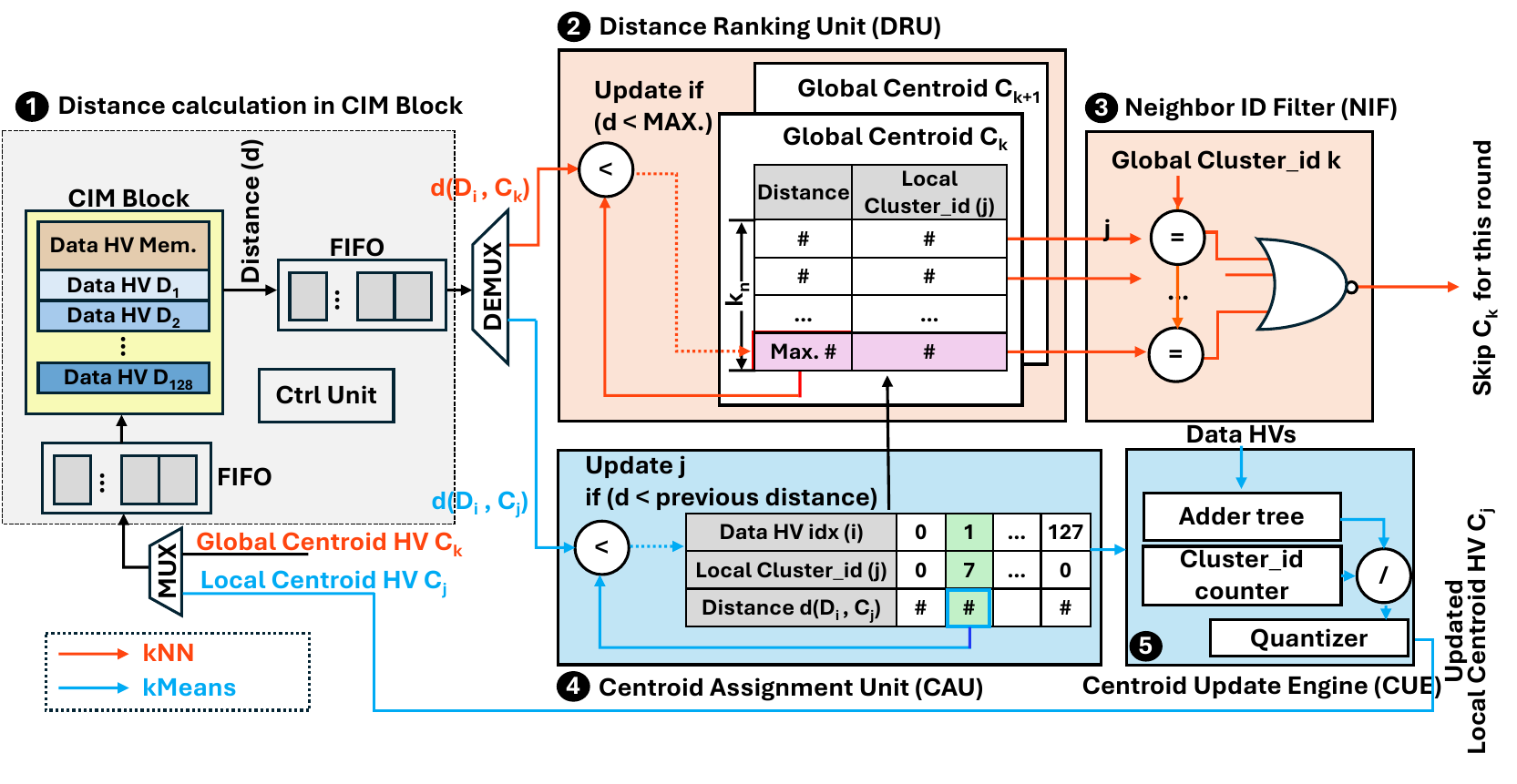}
  \caption{\small Block diagram of the CIM block and digital ASIC modules. DRU and NIF implement kNN-based cluster HV removal, while CAU and CUE support k-means local clustering.
  }
  \label{fig:digital_asic}
  \vspace{-1em}
\end{figure}

\subsection{Digital ASIC Blocks}
\label{digital_ops}
Digital ASIC blocks implement the client-side operations that are not efficiently supported within the CIM blocks, and consume the distance outputs produced by the in-memory computation (Fig.~\ref{fig:digital_asic}, \circled{2}-\circled{5}). After the in-memory distance calculation, the resulting distance values are forwarded to an output FIFO for subsequent digital processing. In our client-update flow, the red path in Fig.~\ref{fig:digital_asic} corresponds to kNN-based HV removal, comprising the CIM block (\circled{1}) and the DRU/NIF blocks (\circled{2}-\circled{3}), and is executed once per communication round for outlier filtering. In contrast, the blue path corresponds to k-means (local clustering) including the CIM block (\circled{1}) and the CAU/CUE blocks (\circled{4}-\circled{5}), and runs in each local iteration to update local centroids.

\noindent\textbf{\circled{2} Distance Ranking Unit (DRU).}
For the kNN-based cluster HV removal phase, DRU is an ASIC module that identifies the $k_n$ nearest data HVs for each centroid by streaming distance values from the CIM blocks. DRU maintains a fixed-size buffer of length $k_n$ that stores the smallest distances and their associated \textit{cluster\_id}s from the previous local clustering round. For each incoming distance value, DRU compares it against the current maximum entry in the buffer and updates the buffer only if the new distance is smaller.

\noindent\textbf{\circled{3} Neighbor ID Filter (NIF).}
NIF is a lightweight comparison unit that receives the selected $k_n$ data HVs and their corresponding \textit{cluster\_ids} from DRU. It checks whether the selected neighbors contain the same \textit{cluster\_id} as the global centroid, and flags the centroid as a local outlier if there is no match through a NOR gate. This outlier flag is then used to skip filtered centroids in the current round.

\noindent\textbf{\circled{4} Centroid Assignment Unit (CAU).}
For the local clustering (k-means) process, CAU compares the newly computed distance for each centroid against the current minimum distance recorded for every data HV. If the new value is smaller, CAU updates both the minimum distance and the associated \textit{cluster\_id}, thereby finalizing the assignment decisions after all centroid HVs have been processed.

\noindent\textbf{\circled{5} Centroid Update Engine (CUE).}
After assignments are determined, CUE processes the assignment results by counting the number of data HVs per cluster using a \textit{cluster\_id} counter and summing the corresponding data HVs with an adder tree. It then computes the averaged HV for each cluster by dividing the adder-tree output by the counter value, and quantizes each element to 3 bits to form the updated centroid HVs. This procedure is repeated for a fixed number of user-specified iterations.

The \Design accelerator instantiates $M+1$ DRUs when the \textit{array row} dimension $M$ is less than or equal to 128, since each DRU can sort 128 distance values in parallel. Similarly, the accelerator incorporates $M$ CAUs, a single NIF, and one CUE.

\begin{figure}[t]
\centering
	  \includegraphics[width=.95\linewidth]{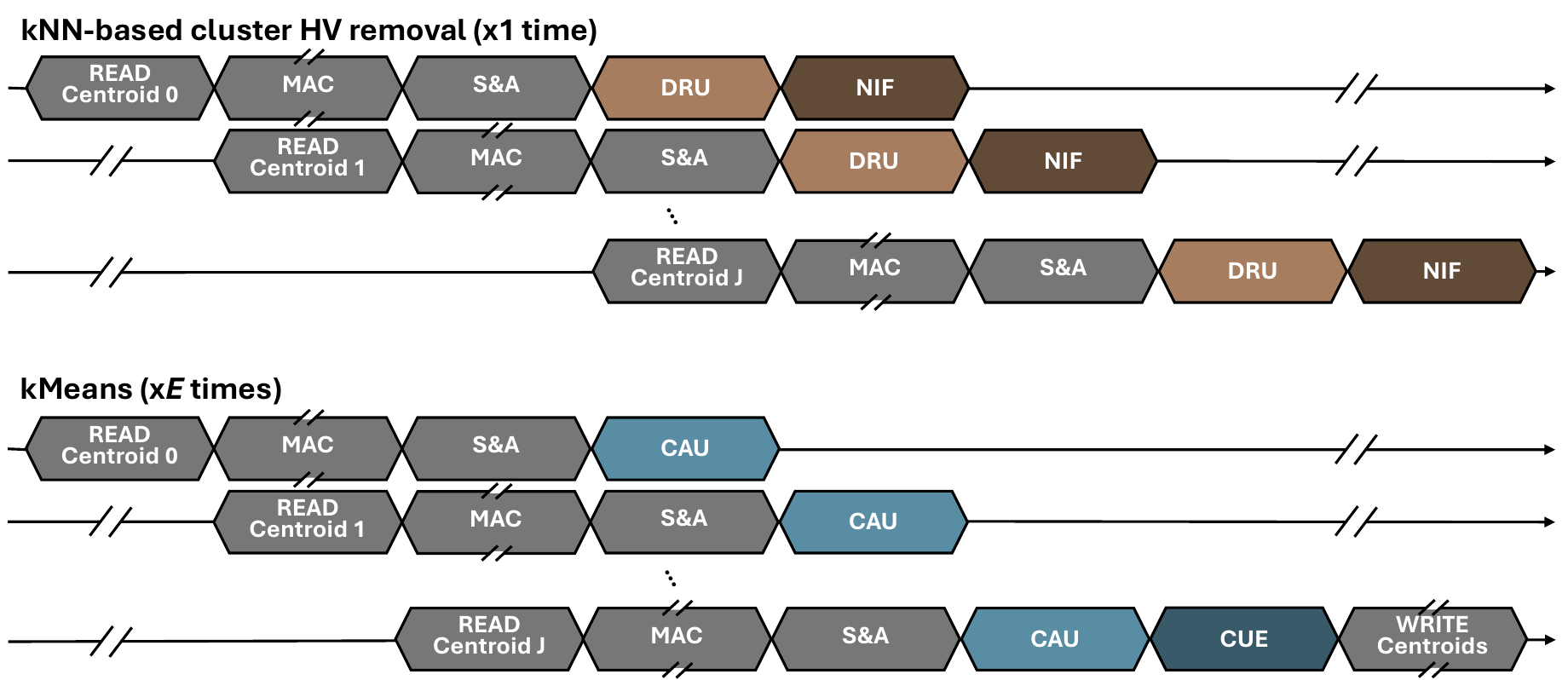}
  \caption{\small Timing diagram of pipelined CIM and digital ASIC operations. Top: kNN-based cluster HV removal. Bottom: k-means local clustering.}
  \label{fig:pipeline}
  \vspace{-1em}
\end{figure}

\subsection{Dataflow and Control Unit}
\label{dataflow}
The overall dataflow of the \Design accelerator is depicted in Figs.~\ref{fig:cim-overview} and~\ref{fig:digital_asic}. The control unit is implemented as a finite state machine that orchestrates data movement across the accelerator and schedules interactions between the CIM blocks and the digital ASIC blocks. The host system first encodes input samples into data HVs and initializes centroid HVs before transferring them to the accelerator. The \Design accelerator stores all data and centroid HVs in the PCM arrays. The encoded data HVs are written once and reused across FL rounds, while the accelerator performs the repeated local clustering updates on chip.

Once initialized, a centroid HV is fetched, buffered in an input FIFO, and converted into an analog input that drives the PCM arrays to compute partial dot products. 
The partial results are then accumulated by per-row S\&A modules to produce final dot-product values, whose negatives are used as distance metrics between data HVs and the active centroid HV. The resulting distance values are written to an output FIFO.
Apart from CIM-based distance computation, all operations in the kNN-based cluster HV removal phase and the k-means (local clustering) process are handled by the four digital ASIC blocks introduced in Sec.~\ref{digital_ops}: DRU, NIF, CAU, and CUE. Alg.~\ref{alg:feduhd} and Fig.~\ref{fig:digital_asic} illustrate two datapaths: (i) kNN-based HV removal and (ii) k-means-based local clustering. The kNN-based HV removal is executed once per round, whereas k-means runs for $E$ iterations in each round.

Figure~\ref{fig:pipeline} illustrates the pipelined execution of the client-side accelerator for (top) kNN-based cluster HV removal (once per round) and (bottom) k-means local clustering ($E$ local epochs). The x-axis denotes time, and each row corresponds to one centroid HV. Each bubble represents a hardware stage; its label indicates the operation being performed, and its width reflects the relative latency (MAC and WRITE latencies are abbreviated). Gray blocks indicate CIM-side operations (read/MAC/S\&A/write), while the colored blocks denote digital ASIC modules: DRU and NIF (brown) for kNN ranking and filtering, and CAU and CUE (blue) for k-means centroid assignment and update. The staggered schedule pipelines consecutive centroids to overlap CIM computation with digital processing. DRU/NIF run for every centroid to generate outlier flags, whereas in k-means CAU runs per centroid and CUE and WRITE are invoked only after the last centroid to update all centroid HVs.

\section{Results}
\subsection{Experimental Setup}

\begin{table}[t]
\centering
\caption{\small Detailed experimental setup of \Design.
\# Instances denotes the number of training samples.
FCL: fully connected layer; Conv: convolutional layer.}
\footnotesize
\begin{adjustbox}{width=\columnwidth}
\begin{tabular}{l|l|l|l|l|l}
\toprule
    & \# Features & \# Instances & Clients & Method   & Baseline \\
    \midrule
    
    HAR~\cite{human_activity_recognition_using_smartphones_240} 
    & 561 & 7,352 & 10 & HDC & 3 FCLs    \\
    
    ISOLET~\cite{cole1991isolet}
    & 617
    & 6,238
    & 10
    & HDC
    & 3 FCLs \\
    
    FMNIST~\cite{xiao2017fashion}
    & $28{\times}28{\times}1$
    & 60,000
    & 10 \& 100
    & HDC
    & 2 Conv + 1 FCL \\
    
    CIFAR10/100~\cite{krizhevsky2009learning}       
    & $32{\times}32{\times}3$ 
    & 50,000 
    & 10 \& 100  
    & FE + HDC 
    & ResNet18 \\
    
\bottomrule
\end{tabular}
\end{adjustbox}
\label{tab:exp_setup}
\vspace{-1em}
\end{table}

\noindent \textbf{Datasets}:
We conduct experiments for various applications, including human activity recognition, speech recognition, and image
classification. 
We use HAR~\cite{human_activity_recognition_using_smartphones_240}, ISOLET~\cite{cole1991isolet}, FMNIST~\cite{xiao2017fashion}, and
CIFAR10/100~\cite{krizhevsky2009learning}, which are standard datasets for these applications. 
Each HAR sample consists of 561 features extracted from IMU sensors, and each ISOLET sample contains 617 acoustic features representing a spoken letter. 
FMNIST consists of $28{\times}28$ grayscale images, while CIFAR10/100 consists of $32{\times}32$ RGB images. 
These datasets are distributed across $I$ clients by sampling class priors from a Dirichlet distribution ($Dir(\alpha)$) with $\alpha=0.1$, following the setting of the state-of-the-art NN-based UFL method~\cite{Orchestra}. 
This yields non-IID, class-imbalanced splits: although the total number of samples per client is balanced, the number of classes per client varies widely (e.g., 4--9 for CIFAR10, 42--50 for CIFAR100, and 2--4 of 6 for HAR, all with 10 clients). 
We evaluate FMNIST and CIFAR10/100 with 10 and 100 clients. HAR and ISOLET are evaluated only with 10 clients, since using 100 clients would leave only approximately 73 and 62 training samples per client, respectively, which is insufficient for stable local unsupervised training and reliable centroid updates.
The detailed setup of \Design on each dataset is summarized in Table~\ref{tab:exp_setup}.

\noindent \textbf{Metrics}:
We use the unsupervised clustering accuracy metric (ACC) that is commonly used in previous unsupervised learning work~\cite{xie2016unsupervised,yu2024lifelong}. Unlike traditional accuracy metrics and the linear evaluation protocols~\cite{Orchestra, FedX}, ACC does not require prior knowledge of the number of classes, nor training a linear classifier with a subset of labeled samples, making it more suitable for unsupervised learning scenarios.
Let $\phi(\theta(\mathbf{x}_j))$ denote the predicted cluster labels and $y_j$ the ground-truth labels.
ACC is computed as follows:
\begin{equation}
\mathrm{ACC}
=
\max_m
\frac{1}{|\mathcal{X}^{\mathrm{test}}|}
\sum_{j=1}^{|\mathcal{X}^{\mathrm{test}}|}
\mathbf{1}
\left\{
y_j =
m\!\left(\phi(\theta(\mathbf{x}_j))\right)
\right\},
\end{equation}
where $m$ ranges over all possible one-to-one mappings between predicted clusters and ground-truth classes. 
Intuitively, ACC measures the accuracy under the ``best'' mapping between the predicted cluster labels and the true labels.
In addition, we examine the training time and energy consumed for end-to-end training.
We also evaluate the communication cost for each communication round.

\noindent \textbf{Baselines}:
We adopt state-of-the-art unsupervised FL frameworks for comparison, including Orchestra~\cite{Orchestra} and its baselines (f-BYOL~\cite{grill2020bootstrap}, f-specloss~\cite{haochen2021provable}, f-simsiam~\cite{chen2021exploring}, f-simclr~\cite{chen2020simple}).
These methods are selected for state-of-the-art performance and realistic settings with a large number of clients (up to 400). 
We exclude FedUL~\cite{FedUL}, FedU~\cite{FedU}, and FedX~\cite{FedX} from our main comparison due to mismatched evaluation settings. Specifically, FedUL assumes shared class distributions across clients~\cite{FedUL}, which is less representative of real-world deployments, while FedU and FedX conduct experiments with a relatively small number of clients (up to 20)~\cite{FedU, FedX}.
For HAR and ISOLET, we use three fully connected layers of 512 units each, while for FMNIST, we use two convolutional layers followed by one fully connected layer, following previous work~\cite{kim2018efficient,FedHD}. 
For CIFAR10/100, we use a ResNet18 backbone~\cite{he2016deep}, following Orchestra~\cite{Orchestra}.
We use Orchestra's original optimization settings unless otherwise noted.

\begin{table}[t]\footnotesize
    \centering
    \caption{\small System specifications of the \Design accelerator. READ/WRITE latency and energy are reported per selected row, while MAC latency and energy are reported per array operation.}
        \begin{tabular}{|c|c|c|}
            \hline
            \multicolumn{3}{|c|}{\textbf{\Design accelerator}} \\ 
            \hline
            \multicolumn{1}{|c|}{Total area} & \multicolumn{2}{|c|}{35.12$\mathrm{mm}^{2}$} \\ 
            \hline
            \multicolumn{1}{|c|}{Array size} & \multicolumn{2}{|c|}{128$\times$128 3-bit PCM cells} \\ 
            \hline
            \multicolumn{1}{|c|}{CIM blocks} & \multicolumn{2}{|c|}{1280 (16$\times$80) arrays} \\
            \hline
            \multicolumn{1}{|c|}{Digital ASIC modules} & \multicolumn{2}{|c|}{S\&A, CAU, DRU, NIF, CUE} \\ 
            \hline
            \hline
             \multicolumn{3}{|c|}{\textbf{Computation Latency and Energy}} \\ 
             \hline
             \multicolumn{3}{|c|}{tREAD=20ns, tWRITE=960ns, tMAC=320ns}\\ 
             \multicolumn{3}{|c|}{eREAD=59.80pJ, eWRITE=2951.04pJ,  eMAC=687.88pJ}\\
             \hline
        \end{tabular}
        \vspace{-1\baselineskip}
    \label{tab:arch_param}
\end{table}

\noindent \textbf{Implementation Details}:
We implement \Design with Python-based libraries~\cite{imambi2021pytorch, marcel2010torchvision}.
We use the standard random projection HDC encoding~\cite{thomas2021theoretical} for HAR, ISOLET, and FMNIST. For CIFAR10/100, we use a ResNet18 backbone~\cite{he2016deep} pretrained on ImageNet~\cite{russakovsky2015imagenet} as a feature extractor before HDC encoding, as shown in Table~\ref{tab:exp_setup}. We emphasize that using a pretrained and frozen feature extractor is standard in HDC-based FL for image datasets~\cite{FHDnn,FedHD}. Since the feature extractor is pretrained on a different dataset, FedUHD has no prior knowledge of the FL workloads at the start of training, ensuring a fair comparison with all baselines.

The number of local epochs ($E$) is set to 10 and the partition ratio ($P$) is 1.0 for all cases. 
We optimize the hypervector dimensions following the prior work~\cite{kim2018efficient, FedHD}, and adopt 1,152 for HAR, 4,096 for ISOLET and FMNIST, and 9,984 for CIFAR10 and CIFAR100. 
These values are chosen to be multiples of 128 to maximize hardware utilization and efficiency on the proposed accelerator. We further analyze the impact of dimensionality on clustering accuracy in Fig.~\ref{fig:algo_analysis}.
The neighborhood size $k_n$ and number of global clusters $J$ are set to $(4,16)$, $(8,64)$, and $(16,128)$ for HAR, ISOLET/FMNIST/CIFAR10, and CIFAR100, respectively, with their values matched to Orchestra's local and global cluster counts~\cite{Orchestra}.
We further perform sensitivity analyses on $k_n$ and $J$, with the results reported in Section~\ref{sec:alg_results}.

\noindent \textbf{Hardware Configurations}:
For \Design CIM accelerator evaluation, we parameterize the PCM primitives, including resistance distribution and read/write latency and energy, using fabricated-device measurements from~\cite{pcm}. Residual device-level non-idealities are modeled by injecting BER during CIM operations, with BER=$10^{-2}$ used by default.
For the CIM peripheral circuits, the schematic and layout are designed in TSMC 40 nm CMOS technology using Cadence Virtuoso, while the digital modules are implemented in Verilog and synthesized with Cadence Genus to obtain area and power.
The architectural parameters are summarized in Table~\ref{tab:arch_param}, and the area breakdown is reported in Fig.~\ref{fig:cim_analysis}(a).
For comparison with alternative CIM substrates, we set up an RRAM-based CIM baseline for \Design using the device characteristics in~\cite{fan2024efficient}. We also include SRAM-CIM performance metrics for NN-based unsupervised learning based on prior SRAM-CIM work~\cite{grimm2022neural, meng2024compute}.

\noindent \textbf{System Environment}:
We evaluate \Design on an NVIDIA H100 GPU and the PCM/RRAM-based CIM accelerators. GPU performance is measured using CUDA utilities and \texttt{nvidia-smi}. Accelerator performance is obtained by integrating the characteristics of all hardware configurations into a NeuroSim~\cite{neurosim}-based Python simulator. The simulator combines these device- and circuit-level primitives following the \Design dataflow to estimate end-to-end latency and energy while accounting for operation dependencies and CIM-array parallelism. For NN-based UFL, GPU performance is measured on the same H100 GPU and SRAM-CIM results are presented with reference to prior SRAM-CIM studies~\cite{grimm2022neural, meng2024compute}.

\begin{table}[t]
\centering
\caption{\small Clustering accuracy (ACC, \%) across all datasets and client settings. 
The highest accuracy is bold, and the second highest is underlined.}
\label{tab:accuracy}
\footnotesize
\begin{adjustbox}{width=\columnwidth}
\begin{tabular}{l|c|c|cc|cc|cc}
\toprule
\multirow{2}{*}{Framework}
& HAR
& ISOLET
& \multicolumn{2}{c|}{FMNIST}
& \multicolumn{2}{c|}{CIFAR10}
& \multicolumn{2}{c}{CIFAR100} \\
& 10
& 10
& 10 & 100
& 10 & 100
& 10 & 100 \\
\midrule
f-BYOL~\cite{grill2020bootstrap} 
& 60.04
& 61.00
& 71.16 & 63.62
& 54.17 & 35.12
& 22.54 & 11.34 \\

f-Specloss~\cite{haochen2021provable}
& 68.60
& 56.38
& 72.35 & 68.57
& 58.99 & 42.30
& \underline{28.85} & 15.33 \\

f-SimSiam~\cite{chen2021exploring} 
& 55.85
& 56.32
& 71.66 & 70.21
& 46.40 & 25.73
& 15.61 & 8.12 \\

f-SimCLR~\cite{chen2020simple}
& 70.77
& 60.48
& 73.63 & \underline{71.16}
& 51.49 & 38.81
& 21.52 & 14.85 \\

Orchestra~\cite{Orchestra} 
& \textbf{76.20}
& \underline{63.28}
& \textbf{76.52} & \textbf{75.79}
& \underline{61.08} & \underline{55.27}
& 25.69 & \underline{18.54} \\

\midrule
\textbf{\Design}
& \underline{74.34}
& \textbf{66.52}
& \underline{74.91} & 69.41
& \textbf{66.40} & \textbf{64.24}
& \textbf{33.59} & \textbf{29.79} \\
\bottomrule
\end{tabular}
\end{adjustbox}
\vspace{-1.0em}
\end{table}

\subsection{\Design algorithm comparison vs. SOTA}
\label{sec:alg_results}

\noindent \textbf{Accuracy}:
Table~\ref{tab:accuracy} reports the clustering accuracy (ACC)~\cite{xie2016unsupervised, yu2024lifelong} of \Design and the NN-based UFL baselines. 
Note that ACC is an unsupervised metric and generally yields lower values than supervised accuracy.
Overall, \Design ranks first or second in seven of the eight evaluated settings, demonstrating competitive representation quality across sensor, speech, and image datasets under non-IID distributions.
On HAR, ISOLET, and FMNIST, where \Design uses only HDC encoding without a pretrained feature extractor, it achieves 74.34\%, 66.52\%, and 74.91\%/69.41\% ACC, respectively. 
In particular, it ranks first or second on all three datasets with 10 clients, showing that its HDC-based learning alone provides competitive representation quality.
On CIFAR10 and CIFAR100, \Design achieves the highest ACC with both client settings, outperforming the strongest baseline by 8.71\%/16.23\% and 16.43\%/60.68\%, respectively, with 10/100 clients. 
The frozen ImageNet-pretrained ResNet18 features alone achieve only 38.85\% and 16.93\% ACC on CIFAR10 and CIFAR100, respectively, when evaluated directly without federated training, indicating that the pretrained features alone do not account for the performance of \Design. 
Together with the results obtained using only HDC encoding and the ablation results in Fig.~\ref{fig:ablation}, these findings support the effectiveness of the kNN-based cluster HV removal and cluster-aware HDC aggregation under non-IID data.

\begin{figure}[t]
\centering
	  \includegraphics[width=\linewidth]{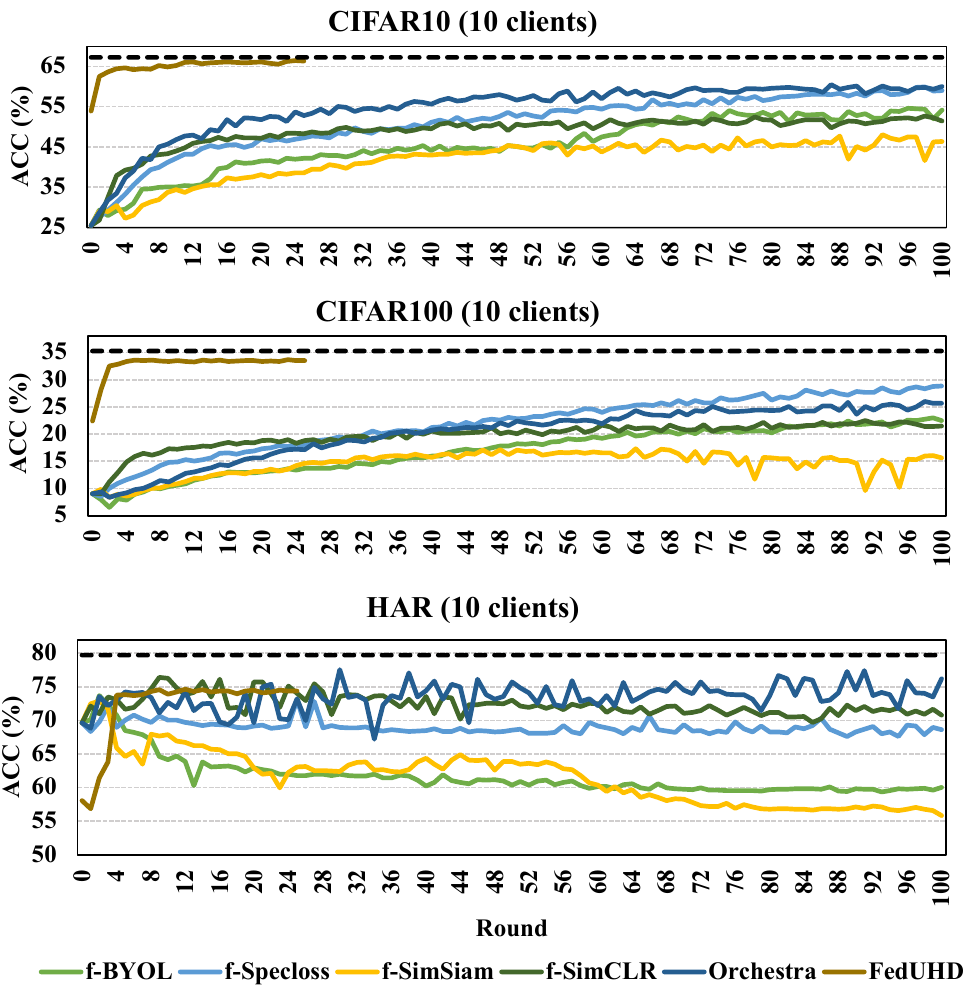}
    \caption{\small Comparison of convergence with NN-based UFL baselines. The x-axis denotes communication round and the y-axis denotes clustering accuracy (ACC). Dashed horizontal lines indicate empirical centralized full-data references obtained using the same $J$, $D$, and 3-bit centroid quantization as \Design.}
    \smallsqueezeup
    \label{fig:convergence}
\end{figure}

\begin{figure}[t]
\centering
	  \includegraphics[width=\columnwidth]{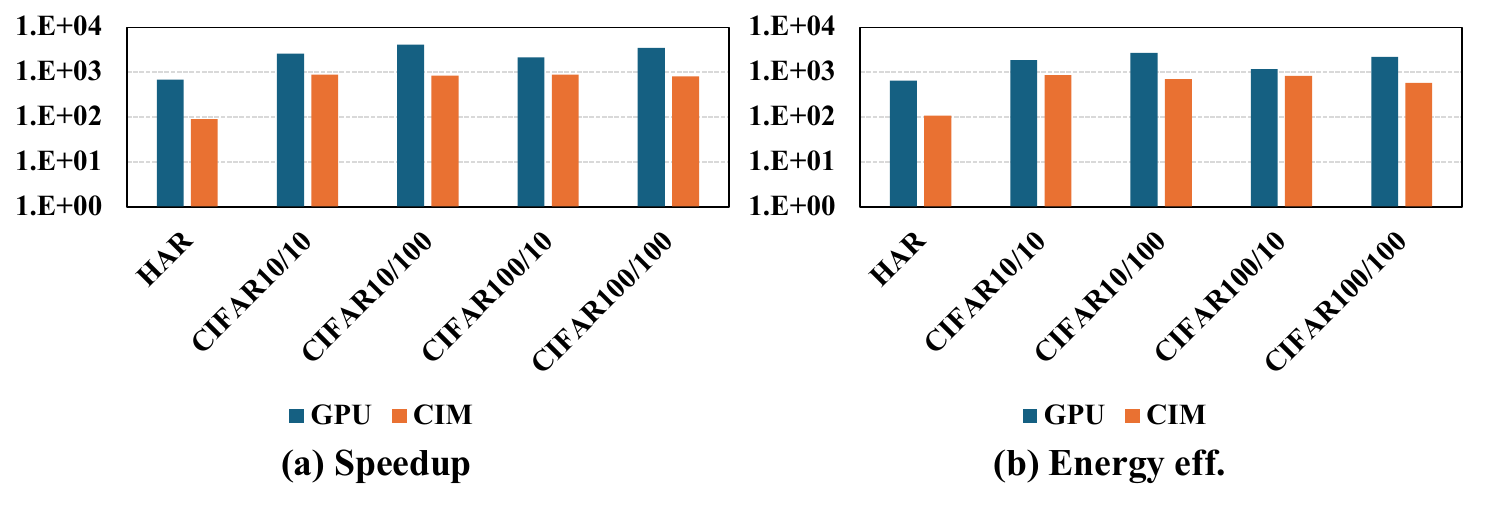}
  \caption{\small Per-client performance comparison between \Design and Orchestra~\cite{Orchestra} on GPU and CIM. Speedup and energy efficiency are shown on a log scale.
  }
  \label{fig:alg_perf}
  \vspace{-1em}
\end{figure}

\noindent
\textbf{Training time and energy efficiency}:
As illustrated in Fig.~\ref{fig:convergence}, one of the key advantages of \Design is its rapid convergence speed. 
Unlike NN-based UFL, which requires intensive computations such as backpropagation for retraining in every round, \Design incurs significantly lower overhead by employing a single kNN operation and a limited number of epochs for k-means clustering.
Another advantage of \Design is that its core HDC operations are highly parallelizable, enabling faster training with a small memory footprint.
In contrast, training NN-based UFL models requires consideration of dependencies across layers, which typically demands memory usage that is approximately three times the number of parameters required for inference~\cite{SAGE}. As a result, the time required for local training is significantly shorter compared to NN-based UFL. Moreover, \Design converges in fewer rounds than NN-based UFL, indicating that the number of communication rounds required for training \Design is considerably lower.

\begin{figure}[t]
\centering
	  \includegraphics[width=\columnwidth]{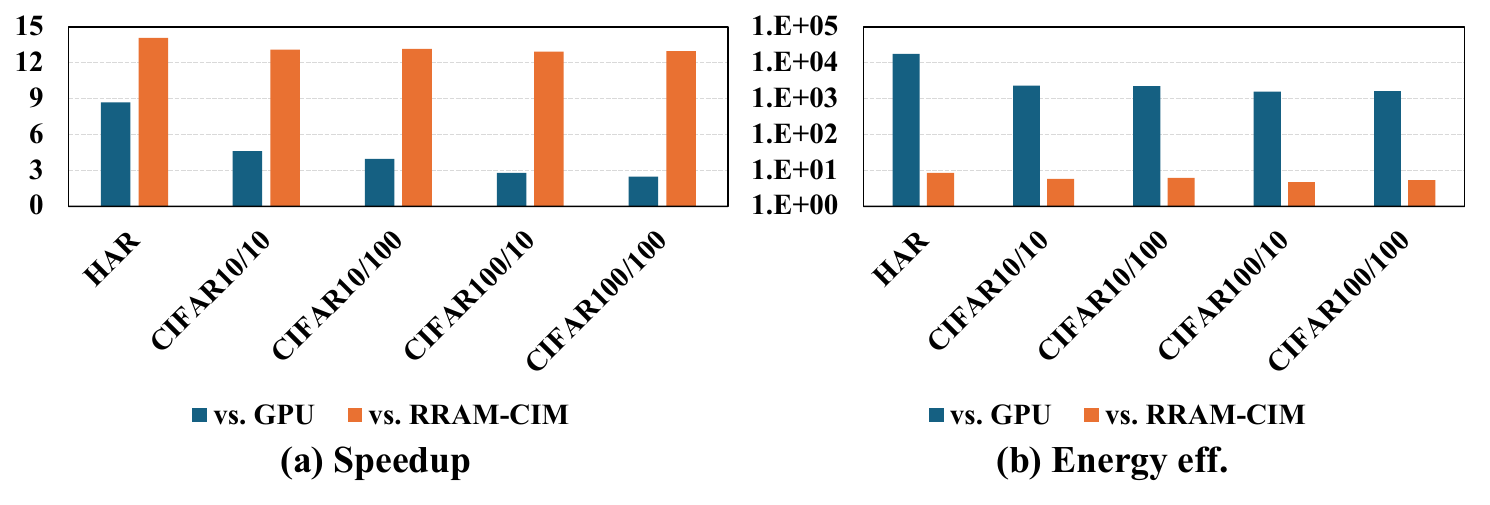}
  \caption{\small Per-client comparison of \Design on PCM-based CIM relative to GPU and RRAM-based CIM. Energy efficiency is shown on a log scale.
  }
  \label{fig:acc_perf}
  \vspace{-1em}
\end{figure}

Fig.~\ref{fig:alg_perf} highlights the benefit of the \Design algorithm itself in terms of per-client training speed and energy efficiency, while the hardware-specific benefit of our PCM-based CIM design is analyzed separately later in Fig.~\ref{fig:acc_perf}. This algorithmic efficiency can be more clearly understood by examining the required computation and memory footprint. For example, when training 1,000 32$\times$32 images at a client for 10 epochs, a ResNet18-based UFL method (e.g., Orchestra~\cite{Orchestra}) without fully connected layers requires 471.2 GFLOPs of total floating-point operations per client and a 187.9 MB memory footprint with a batch size of 128. In contrast, \Design in full precision requires only 2.076 GFLOPs and a 40.45 MB memory footprint on average by applying kNN-based cluster hypervector removal. For the GPU results in Fig.~\ref{fig:alg_perf}, \Design~(including encoding) and Orchestra are evaluated on an NVIDIA H100 GPU. For the CIM evaluation, \Design is analyzed using a hybrid setup: the encoding stage is modeled based on prior SRAM-CIM NN accelerator~\cite{grimm2022neural} and average CIM power~\cite{meng2024compute}, whereas the local clustering stage is executed on our PCM-based CIM accelerator. For the CIM-based evaluation of Orchestra, end-to-end training is analyzed using the same NN-oriented CIM model~\cite{grimm2022neural} and power model~\cite{meng2024compute}. 
Averaged across the five hardware-evaluation workloads in Fig.~\ref{fig:alg_perf}, \Design achieves three orders of magnitude higher speedup (2,239$\times$) and energy efficiency (1,542$\times$) than Orchestra on GPU, and two orders of magnitude improvements (549$\times$ for speedup and 502$\times$ for energy efficiency) on CIM.
The smaller gap on CIM arises because the HDC-based clustering stage of \Design is highly amenable to PCM-CIM acceleration, leaving the encoding stage as the dominant bottleneck.

\noindent
\textbf{Communication cost}:
Fig.~\ref{fig:convergence} shows that \Design achieves accuracy convergence in fewer rounds so that it completes training much earlier than NN-based UFL. We stop \Design after 25 rounds as the accuracy saturates.
Additionally, \Design incurs lower communication overhead than NN-based UFL in each round. In NN-based UFL, the feature extractor must be exchanged every round. In contrast, \Design only requires the transmission of local cluster HVs and the size of each cluster, significantly reducing the data exchanged between the server and clients each round. In detail, \Design needs to transfer the product of the number of clusters 16/64/128 trained on HAR/CIFAR10/100 and the dimension of each local cluster HV (1,152 for HAR and 9,984 for CIFAR10/100) with just 25 rounds. This yields 176$\times$/73$\times$/37$\times$ lower communication costs than the baselines on HAR/CIFAR10/CIFAR100, respectively.

\begin{figure}[t]
\centering
    \includegraphics[width=\columnwidth]{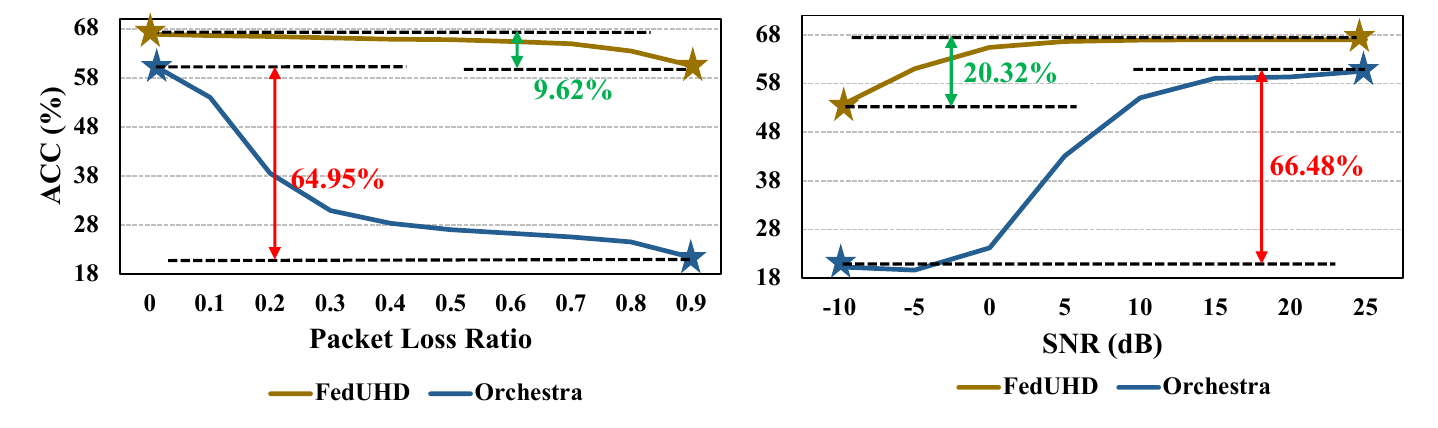}
    \caption{\small Comparison of robustness with NN-based UFL~\cite{Orchestra} on the CIFAR10 dataset with 10 clients.}
    \smallsqueezeup
    \label{fig:robustness}
    \vspace{-1em}
\end{figure}

\noindent
\textbf{Robustness}:
Robustness is a critical factor due to the unreliable communication encountered by edge devices. As illustrated in Fig.~\ref{fig:robustness}, we evaluate two different communication failure scenarios: packet loss and the addition of Gaussian noise, both of which are commonly encountered in real-world AIoT transmission channels~\cite{FHDnn}. Note that robustness is evaluated using binary data HVs and 3-bit centroid HVs. The results demonstrate that \Design exhibits superior robustness across all scenarios when compared to Orchestra~\cite{Orchestra}. For quantitative analysis, accuracy degradation is calculated as follows:
\vspace{2mm}
\begin{equation}
\begin{split}
\frac{(base~accuracy) - (perturbed~accuracy)}{base~accuracy} *100
\end{split}
\end{equation}

\noindent Under packet loss conditions, the accuracy degradation is 9.62\% for \Design and 64.95\% for Orchestra, while under Gaussian noise conditions, the accuracy degradation is 20.32\% for \Design and 66.48\% for Orchestra. These results show that the HV-based model of \Design exhibits greater resilience to noise compared to the state-of-the-art NN-based UFL method~\cite{Orchestra}.

\begin{figure}[t]
\centering
    \includegraphics[width=\columnwidth]{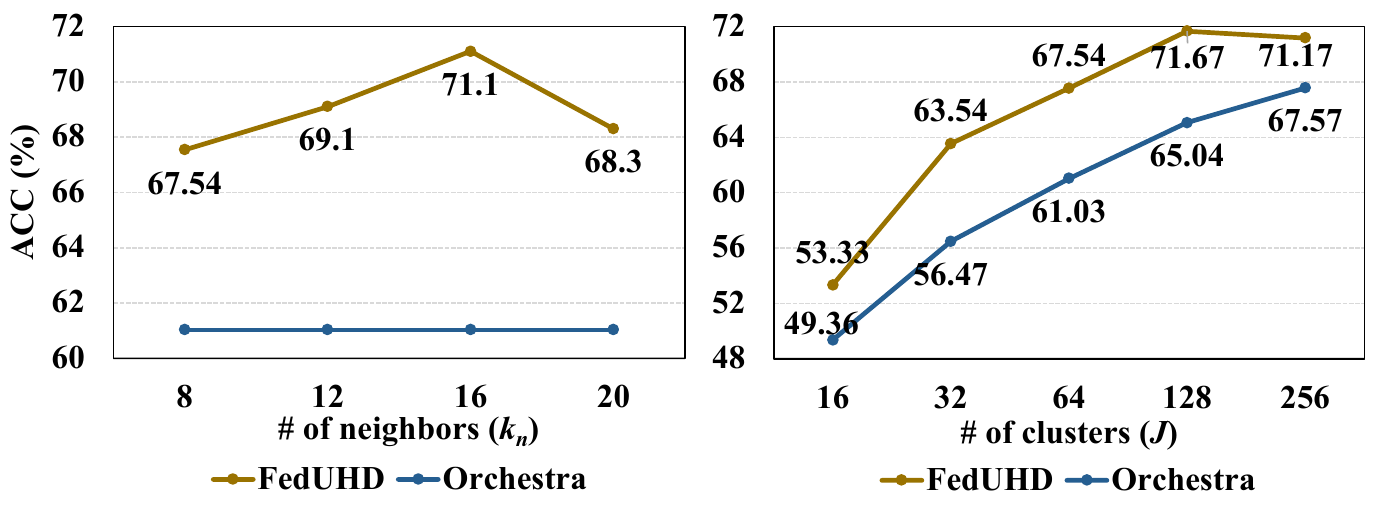}
    \caption{\small Final ACC under different neighborhood sizes $k_n$ and cluster counts $J$ on CIFAR10 with 10 clients.}
    \smallsqueezeup
    \label{fig:sensitivity}
\end{figure}

\noindent\textbf{Sensitivity Analysis}:
Fig.~\ref{fig:sensitivity} illustrates the impact of $k_n$ as well as the number of clusters $J$ in \Design and Orchestra~\cite{Orchestra}. Sensitivity analysis is also conducted using full-precision data HVs and centroid HVs to validate feasibility.
$k_n$ adjusts the neighborhood size in the kNN-based HV removal phase, with $k_n=16$ generating the best ACC.
Overall, \Design shows robust performance with $k_n$ in a reasonable range between 8 and 20, consistently outperforming Orchestra.
The number of clusters $J$ is a key design choice in cluster-based UFL methods such as \Design and Orchestra when the number of ground-truth classes is not provided.
As shown in Fig.~\ref{fig:sensitivity} (right), increasing $J$ improves ACC by allowing finer-grained clusters to align better with true labels. Notably, for the same number of global clusters, \Design consistently outperforms Orchestra. Because the kNN-based removal prunes cluster HVs lacking local support, \Design can tolerate an over-provisioned $J$, so $J$ need not match the unknown class count exactly within the evaluated range. Empirically, this removal reduces the active clusters retained per client to a smaller, data-supported subset according to each client's local data support.

\subsection{\Design Algorithm Analysis}
\begin{figure}[t]
\centering
	  \includegraphics[width=\columnwidth]{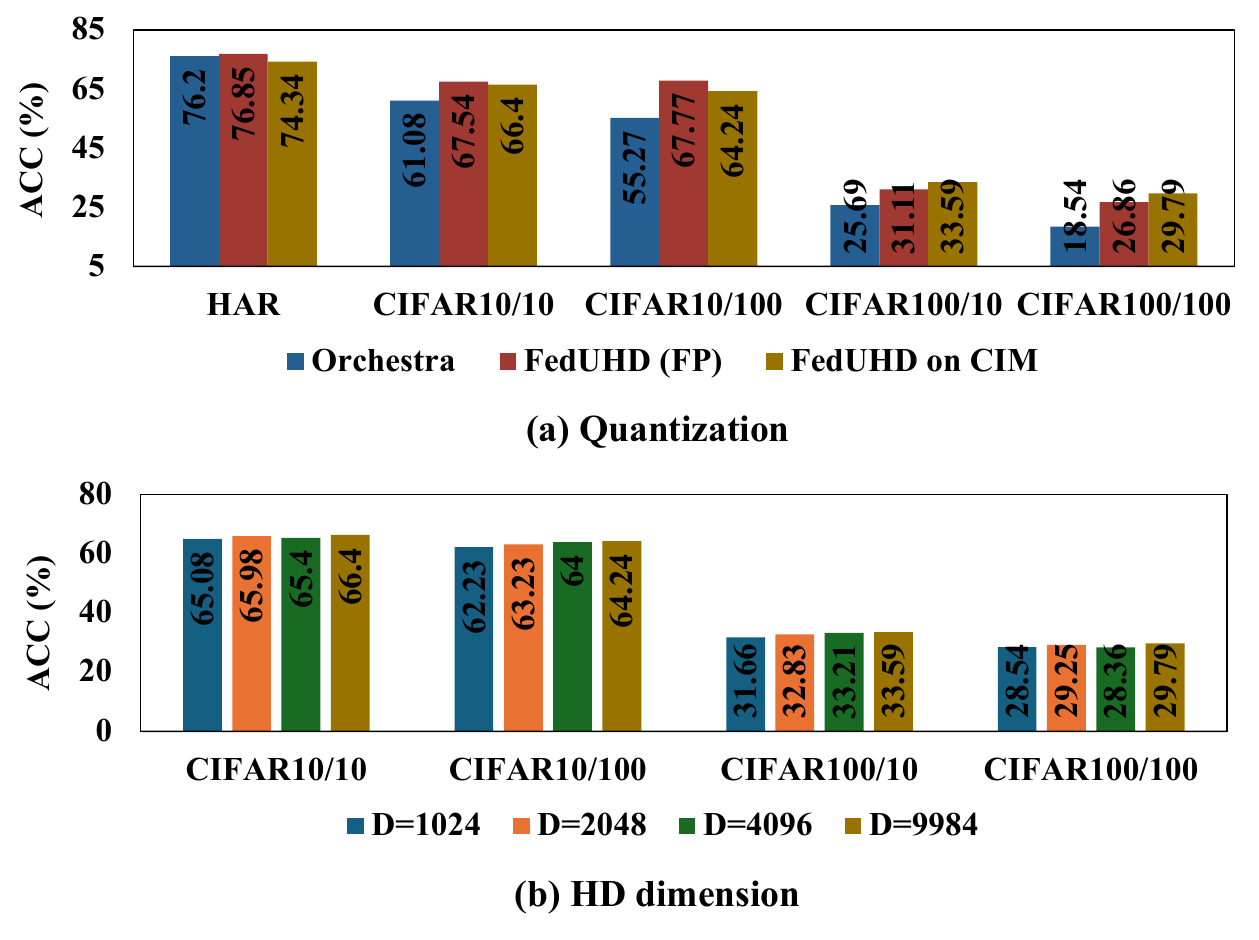}
  \caption{\small \Design algorithm analysis: (a) effect of quantization (b) relationship between HD dimension and ACC
  }
  \label{fig:algo_analysis}
\vspace{-1em}
\end{figure}

\noindent\textbf{Effect of quantization on accuracy}: 
Fig.~\ref{fig:algo_analysis}(a) compares clustering accuracy across Orchestra~\cite{Orchestra}, \Design with full precision (FP), and \Design on CIM. Although the CIM implementation uses binary data HVs, 3-bit centroid HVs, and random bit flips during in-memory operations, \Design on CIM achieves accuracy comparable to the full-precision baseline. This result stems from the inherent resilience of HDC.
Based on internal evaluations, \Design maintains comparable clustering accuracy with ADC resolutions down to 5-bit for partial sum precision in the CIM block, while accuracy degrades severely at 4-bit ADC resolution.

\noindent\textbf{Effect of memory bit errors}: 
We evaluate the robustness of \Design on CIM by injecting random bit errors into all CIM operations. In our bit-error rate (BER) sweep, \Design achieves 66.40\% accuracy at the default BER of $10^{-2}$ and maintains comparable accuracy up to a BER of $0.25$, while accuracy drops to 36.75\% and 16.77\% at BERs of $0.33$ and $0.50$, respectively. These results are measured on CIFAR10 with 10 clients.

\noindent\textbf{HD dimension vs ACC}:
The HD dimension introduces a trade-off between accuracy and hardware energy consumption. In supervised settings, accuracy increases with the HD dimension but eventually saturates while energy cost continues to grow~\cite{relHD}. A similar trend is observed in the unsupervised setting. 
As shown in Fig.~\ref{fig:algo_analysis}(b), accuracy on CIFAR10/100 saturates near $D=9{,}984$, which we adopt for these datasets.
This choice is consistent with the HD dimension of 10,000 commonly adopted in prior work on supervised federated learning with HDC~\cite{FedHD}.

\begin{figure}[t]
\centering
	  \includegraphics[width=\columnwidth]{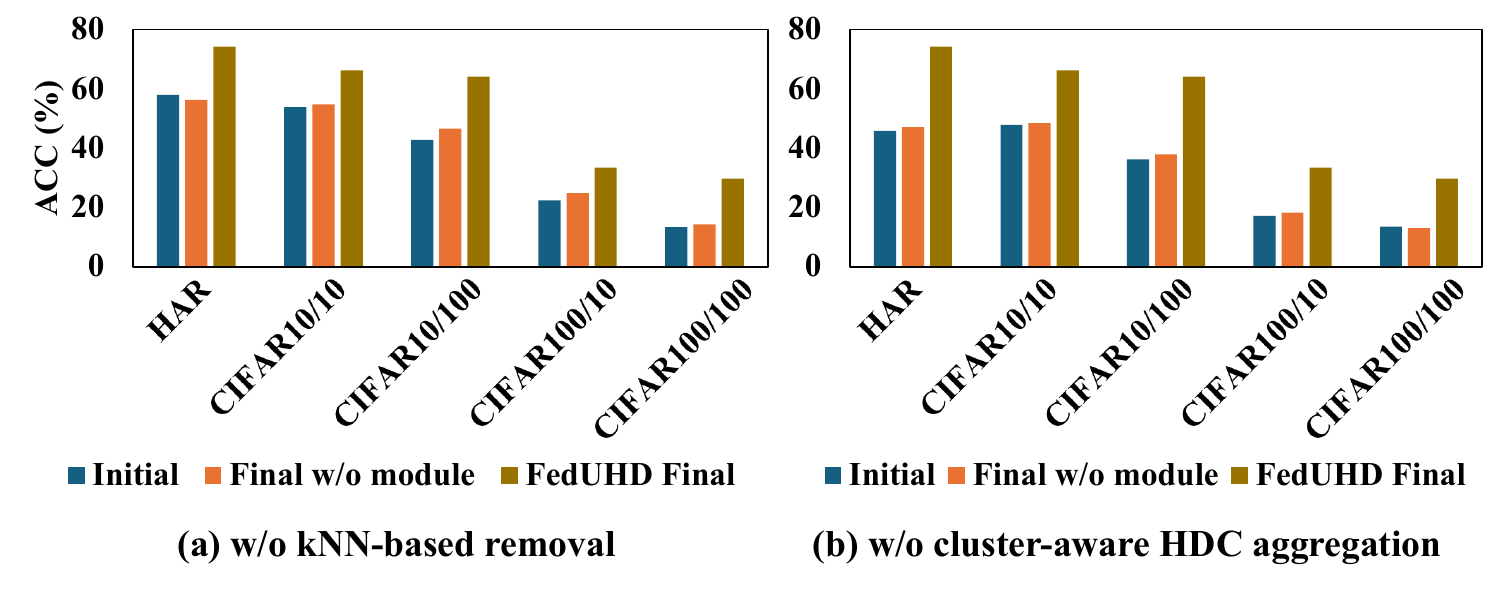}
  \caption{\small \Design ablation study: (a) effect of the kNN-based cluster HV removal (b) effect of the cluster-aware HDC aggregation
  }
  \label{fig:ablation}
  \vspace{-1em}
\end{figure}

\noindent\textbf{Ablation study of key algorithmic components}:
Fig.~\ref{fig:ablation} evaluates the contribution of the two key algorithmic components in \Design: the kNN-based cluster HV removal and the cluster-aware HDC aggregation. In Fig.~\ref{fig:ablation}(a), removing the kNN-based cluster HV removal means that all global centroid HVs are directly assigned to the initial centroid HVs at each communication round without filtering. Under this setting, accuracy even drops below the initial value for HAR and CIFAR10 with 10 clients, and although some other settings still improve over the initial state, they remain consistently below the full \Design. This result indicates that the kNN-based module is critical for filtering unhelpful global centroids and stabilizing local training, yielding an average accuracy gain of 39.4\% across the evaluated ablation settings.

In Fig.~\ref{fig:ablation}(b), removing the cluster-aware HDC aggregation replaces the server update with uniform averaging of local centroids from all clients at each round. While most settings still outperform the initial accuracy, except for CIFAR100 with 100 clients, the performance is again consistently lower than that of the full \Design. This confirms that client-side refinement alone is insufficient under non-IID data distributions, and that server-side aggregation must also account for centroid-level heterogeneity. Overall, the cluster-aware HDC aggregation provides a 68.8\% average accuracy improvement over uniform aggregation across the evaluated ablation settings.

\subsection{\Design Accelerator vs. \Design on GPU}
Fig.~\ref{fig:acc_perf} compares training latency and energy efficiency between the proposed \Design PCM-based CIM accelerator and a GPU baseline. Unlike the algorithm-level comparison in Sec.~\ref{sec:alg_results}, this accelerator-level comparison excludes the common one-time host-side encoding step and compares only the repeated local training pipeline after encoded HVs are available. Across the five hardware-evaluation workloads in Fig.~\ref{fig:acc_perf}, the \Design accelerator consistently outperforms the GPU. Averaged over datasets and client configurations, the CIM design achieves a $4.07\times$ speedup and a $2{,}981\times$ improvement in energy efficiency. 
These gains stem from executing distance computation directly inside PCM crossbar arrays which minimizes data movement and avoids the memory bandwidth bottleneck inherent to GPU-based HDC training. In addition, dedicated digital modules handle kNN-based filtering and centroid updates locally, which further reduces off-chip memory accesses. 
We observe that the benefit of minimizing data movement affects energy efficiency much more strongly than speedup. This trend is supported by comparison with another CIM-based design in Fig.~\ref{fig:acc_perf}, where the relative energy efficiency gain is only $6.04\times$ on average.

The per-workload trends in Fig.~\ref{fig:acc_perf} are mainly explained by workload size, GPU overhead, and write intensity. HAR shows larger gains because its smaller dimensionality reduces activated CIM blocks, while GPU execution is dominated by fixed overhead and baseline power. As the number of clients increases, the per-client dataset size decreases, reducing GPU latency but not GPU power proportionally. Moving from CIFAR10 to CIFAR100 increases the number of clusters and centroid-update writes, which reduces the relative speedup because PCM writes are more expensive than reads.

\subsection{\Design on PCM-CIM vs. \Design on RRAM-CIM}
Fig.~\ref{fig:acc_perf} compares the per-client training latency and energy efficiency of \Design on PCM-based and RRAM-based CIM~\cite{fan2024efficient}, using the same digital ASIC modules for both platforms.
Across the five hardware-evaluation workloads, PCM-based CIM consistently outperforms RRAM-based CIM, achieving average gains of $13.22\times$ in speed and $6.04\times$ in energy efficiency.
These gains stem from PCM's lower write latency and energy, which reduce the cost of repeated centroid updates during local clustering.
Accordingly, the gains are larger for write-intensive workloads, such as HAR and settings with fewer samples per client, and smaller when MAC operations dominate.

\begin{figure}[t]
\centering
	  \includegraphics[width=\columnwidth]{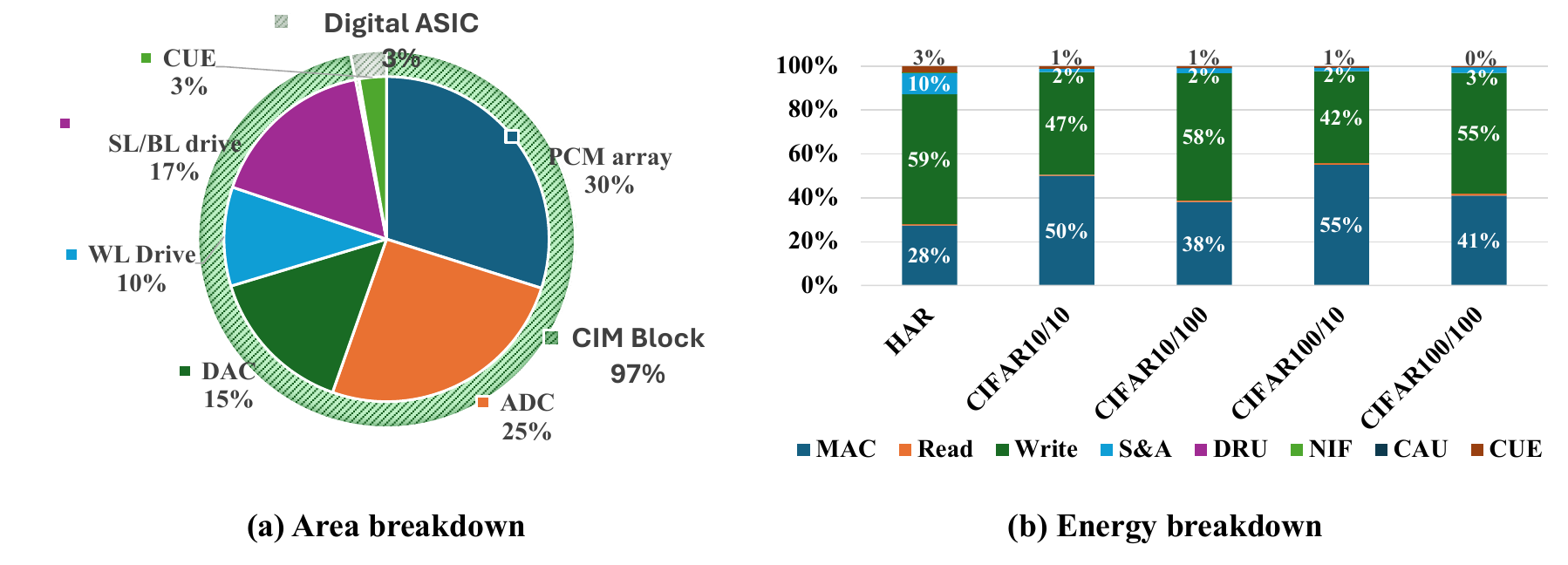}
  \caption{\small \Design accelerator analysis: (a) area breakdown (b) energy breakdown.
  }
  \label{fig:cim_analysis}
  \vspace{-1em}
\end{figure}

\subsection{\Design Accelerator Analysis}
\label{CIM_analysis}
\noindent\textbf{Area Breakdown}:
Fig.~\ref{fig:cim_analysis}(a) illustrates the area breakdown of the CIM block and the overall \Design accelerator. 
Within the CIM block, the PCM array accounts for the largest area fraction, followed by the ADC among peripheral circuits. We adopt 16 row-multiplexed 5-bit flash ADCs per array, with each ADC shared across 8 rows through an 8:1 multiplexer, which provides a favorable tradeoff between algorithmic stability and hardware cost. We observe that lower-resolution ADCs lead to training failure in \Design, yet the ADC area overhead of 25\% remains acceptable given that prior work reports ADCs accounting for 35\%--60\% of total area~\cite{fan2024specpcm, kim2022processing}.
Other peripheral components, including the WL driver, SL/BL driver, and DACs, account for 10\%, 17\%, and 15\% of the area, respectively. At the accelerator level, the digital ASIC submodules account for only 3\% of the total area, indicating that the additional logic for kNN-based filtering and local clustering introduces marginal overhead compared to the CIM blocks.

\noindent\textbf{Energy Breakdown}:
Fig.~\ref{fig:cim_analysis}(b) shows that MAC and write operations account for approximately 48\% and 49\% of the energy, respectively, across the five hardware-evaluation workloads, while the digital ASIC modules account for only approximately 3\%.
The relative MAC and write contributions are mainly determined by the HD dimension ($D$), local sample count, and cluster count ($J$), since MAC operations are performed for every sample--centroid comparison, whereas centroid HVs are written once per iteration.
Accordingly, HAR has lower MAC and higher write fractions than CIFAR (28\%/59\% versus 48\%/49\% on average), and increasing the number of clients from 10 to 100 shifts the average fractions from 53\%/44\% to 40\%/56\%.
Moving from CIFAR10 to CIFAR100 increases the average MAC fraction from 45\% to 49\% and decreases the write fraction from 51\% to 47\%, reflecting the larger cluster count.

\noindent\textbf{Endurance analysis}:
We also analyze PCM endurance for the proposed training workload. The adopted superlattice PCM device demonstrates endurance beyond $10^8$ switching cycles~\cite{pcm}. In \Design, data HVs are written once when loaded and then reused through read/MAC operations, while centroid HVs are rewritten only during centroid-update steps. With $E=10$ local iterations and $R=25$ training rounds, each centroid cell experiences approximately $E\times R=250$ write updates per training run, which is several orders of magnitude below the demonstrated PCM endurance.

\section{Conclusion}
We proposed \Design, the first HDC-based UFL framework, combining client-side kNN-based cluster HV removal, server-side cluster-aware HDC aggregation, and a PCM-based CIM accelerator.
Across all datasets, \Design achieves competitive accuracy against state-of-the-art NN-based UFL methods.
On HAR and CIFAR10/100, it provides average speedup and energy-efficiency gains of 2,239$\times$/1,542$\times$ on GPU and 549$\times$/502$\times$ on CIM, while reducing communication cost by up to 176$\times$ and demonstrating greater robustness than Orchestra under communication noise.
Compared with GPU execution of \Design, the proposed accelerator further provides a 4.07$\times$ speedup and 2,981$\times$ higher energy efficiency, while outperforming RRAM-based CIM by 13.22$\times$ and 6.04$\times$, respectively.

\section*{Acknowledgment}
This work was supported in part by CoCoSys and PRISM, centers in JUMP 2.0, an SRC program sponsored by DARPA, SRC Global Research Collaboration (GRC) grants, and NSF grants \#1826967, \#1911095, \#2003279, \#2052809, \#2112665, \#2112167, \#2100237, and \#2211386.

%\smallsqueezeup
% \setstretch{0.9}
\bibliographystyle{IEEEtran}
\bibliography{references}

\end{document}